\documentclass{article}

\usepackage[preprint]{corl_2026} 

\usepackage{url}
\usepackage[utf8]{inputenc}
\usepackage{graphicx}
\usepackage{amsmath}
\usepackage{amsthm}
\usepackage{booktabs}
\usepackage{algorithm}
\usepackage{algorithmic}

\usepackage{multirow}
\usepackage{booktabs}
\usepackage{array}
\usepackage{amsmath}
\usepackage{mathtools}

\usepackage{amssymb}
\usepackage{pifont}

\usepackage{algorithm}
\usepackage{algorithmic}
\usepackage{placeins}
\usepackage{subcaption}
\usepackage{makecell}
\usepackage[inline]{enumitem}

\usepackage[normalem]{ulem}
\useunder{\uline}{\ul}{}
\usepackage{wrapfig}

\definecolor{red}{rgb}{0.9, 0.1, 0.1}

\usepackage{fontawesome5}

\newcommand{\todo}[1]{{{}}}

\definecolor{darkgreen}{rgb}{0.0, 0.55, 0.0}
\newcommand{\cmark}{{\color{darkgreen}\ding{51}}}     
\newcommand{\xmark}{{\color{red}\ding{55}}}
\definecolor{gray}{rgb}{0.1, 0.1, 0.1}
\newcommand{\nmark}{{\color{gray}{n/a}}}

\newcommand{\RBM}{RoboBenchMart}
\usepackage{xspace}

\title{RoboBenchMart: Benchmarking Robots in Retail Environment}

\newcommand{\weburl}{\url{https://github.com/emb-ai/RoboBenchMart}\xspace}

\author{%
\textbf{Konstantin Soshin}\textsuperscript{$1*$} \quad
\textbf{Alexander Krapukhin}\textsuperscript{$1*$} \quad
\textbf{Andrei Spiridonov}\textsuperscript{$1*$} \\ 
\textbf{Gregorii Bukhtuev}\textsuperscript{$1$} \quad
\textbf{Andrey Kuznetsov}\textsuperscript{$1$} \quad
\textbf{Vlad Shakhuro}\textsuperscript{$1,2,3$} \quad
\textbf{Denis Shepelev}\textsuperscript{$1,2$$\dagger$} \\
\textsuperscript{$1$}FusionBrain Lab, Robotics Group\quad
\textsuperscript{$2$}NUST MISIS\quad
\textsuperscript{$3$}Lomonosov Moscow State University\\
 \textsuperscript{$*$}Equal contribution\quad
 \textsuperscript{$\dagger$}Project leader\quad\\
 \faGithub\;\weburl
}

\begin{document}

\maketitle

\vspace{-1em}

\begin{abstract}
Most existing robotic manipulation benchmarks focus on tabletop or household scenarios.
While these setups have driven impressive progress, it remains unclear whether generalist VLAs that excel there can truly generalize to domains with different geometry, semantics, and workflows.
We introduce \textbf{RoboBenchMart}, an open-source simulated benchmark targeting retail dark-store environments, where a mobile manipulator must perform complex manipulation tasks with diverse grocery items.
This setting presents significant challenges, including dense object clutter and varied spatial configurations, with items positioned at different heights, depths, and in close proximity.
By targeting on the retail domain, our benchmark addresses a setting with strong potential for near-term automation impact.
Using generated trajectories, we model a standard, realistic fine-tuning setup for current generalist VLAs and evaluate several state-of-the-art models.
We find that they still struggle even on common retail tasks, indicating that these models are not yet truly general across domains.
To support further research, we release the RoboBenchMart suite, which includes a procedural store layout generator, a trajectory generation pipeline, evaluation tools, and fine-tuned baseline models.
\end{abstract}


\section{Introduction}

\begin{wrapfigure}{r}{0.5\textwidth}
    \vspace{-1em}
    \centering
    \includegraphics[width=\linewidth]{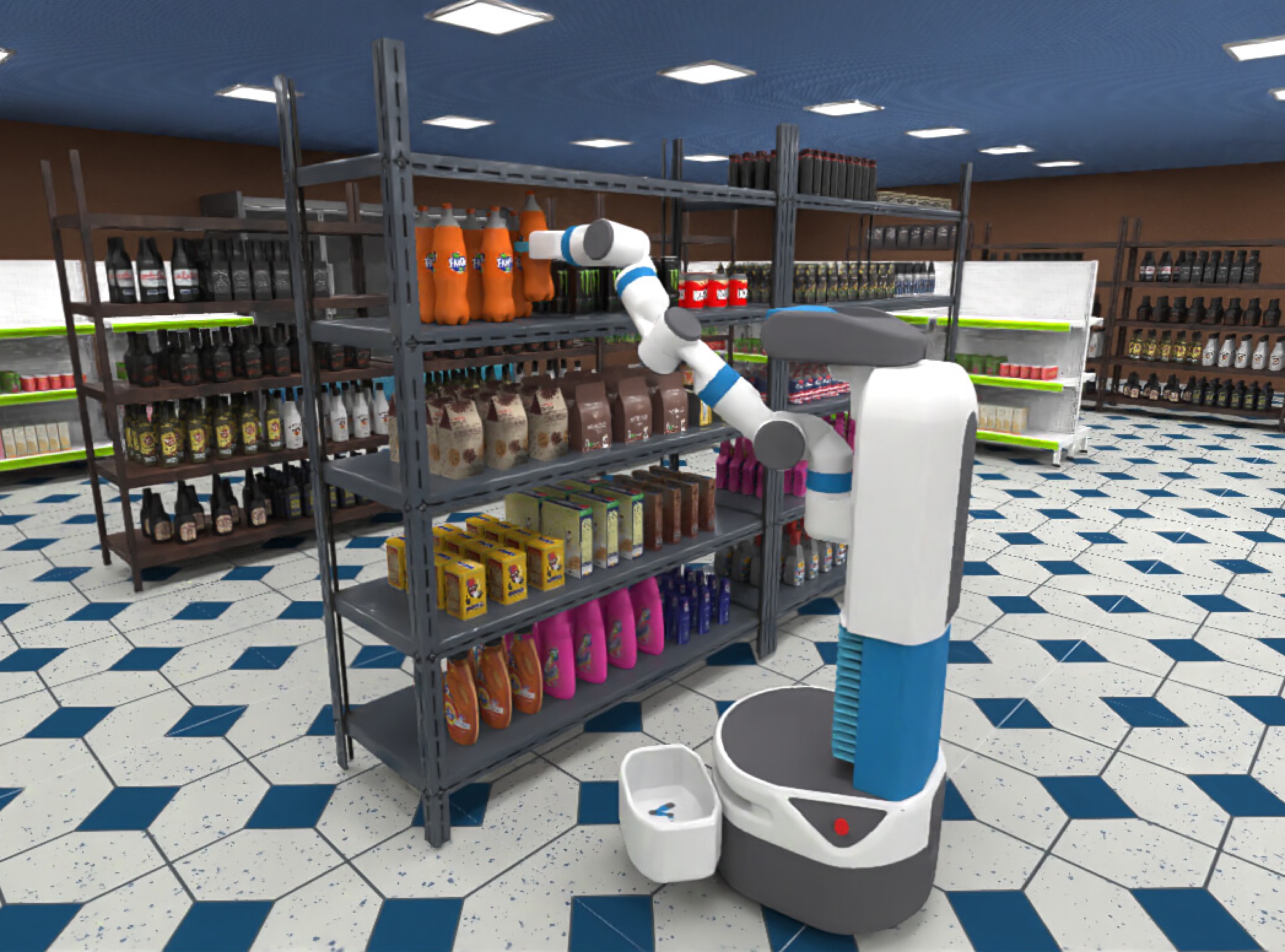}
    \caption{\RBM{} in action\,---\,the Fetch robot operates in a realistic, cluttered retail environment.}
    \label{fig:teaser}
    \vspace{-2em}
\end{wrapfigure}
Current robotic systems deployed in the real world mainly operate in constrained environments and typically perform a single, specific task.
While these systems provide significant economic value, the long-term goal of robotics is to develop systems that can operate in unconstrained, noisy, realistic settings, generalize across a wide range of variations, and perform multiple tasks.
Recent advances in multimodal deep learning and robotics bring us closer to this goal, but widespread deployment in unstructured environments remains distant.

A large body of recent work addresses this gap through simulated benchmarks for manipulation and navigation.
However, most benchmarks used to develop and evaluate ``generalist'' VLAs focus on tabletop or household scenarios (see Table~\ref{tab:review:bench}), where most tasks can be reduced to pick-and-place.
Strong performance in these settings does not guarantee that a policy will transfer or adapt to domains with different geometry, semantics, and workflows.
This motivates the development of \emph{domain-specific} simulation benchmarks that expose qualitatively different challenges while still enabling systematic evaluation.

One such promising domain is retail.
Dark stores, in particular, have gained popularity worldwide as small retail distribution centers that fulfill online orders.
They are not open to the public; instead, workers (and potentially robots) perform restocking and order picking.
We view automation of such tasks as a realistic near-term goal: dark stores involve limited human–robot interaction and are more structured than conventional grocery stores. 
Automating repetitive picking and restocking can reduce physical strain and injury risk for workers, enable 24/7 operation, and improve the efficiency and reliability of order fulfillment, ultimately benefiting customers through more stable and accessible delivery services.

\begin{table*}[!hbt]
\centering
\resizebox{\columnwidth}{!}{%
\fontsize{9pt}{10pt}\selectfont
\tabcolsep=2.5pt
\begin{tabular}{cccccccccc}
\toprule
\begin{tabular}[c]{@{}l@{}}Benchmark/\\ Dataset\end{tabular} &
\begin{tabular}[c]{@{}c@{}}Published \\ in\end{tabular} &
  \begin{tabular}[c]{@{}c@{}}Retail \\ Domain\end{tabular} &
  \begin{tabular}[c]{@{}c@{}}Scene \\ Generation\end{tabular} &
  \begin{tabular}[c]{@{}c@{}}Arrangement \\ Generation\end{tabular} &
  \begin{tabular}[c]{@{}c@{}}Release \\ 3D Assets\end{tabular} &
  \begin{tabular}[c]{@{}c@{}}Trajectories \\ Generation\end{tabular} &
  \begin{tabular}[c]{@{}c@{}}Tasks \\ Diversity\end{tabular} &
  \begin{tabular}[c]{@{}c@{}}Atomic \\ Tasks\end{tabular} &
  \begin{tabular}[c]{@{}c@{}}Composit. \\ Tasks\end{tabular} \\
\midrule
ALFRED & CVPR'20 & \xmark & \cmark      & \cmark & \cmark & \cmark & \cmark            & \cmark & \cmark \\
RLBench & RA-L'19 & \xmark & \xmark      & \cmark & \cmark & \cmark & \cmark            & \cmark & \cmark \\
RoboCasa & RSS'24 & \xmark & \cmark      & \cmark & \cmark & \cmark & \cmark            & \cmark & \cmark \\
CALVIN & RA-L'22  & \xmark & \xmark      & \cmark & \nmark & \cmark & \cmark            & \cmark & \cmark \\
LIBERO & NeurIPS'23   & \xmark & \cmark      & \cmark & \cmark & \cmark & \cmark            & \cmark & \cmark \\
VLABench & arXiv'24   & \xmark & \cmark      & \cmark & \cmark & \cmark & \cmark            & \cmark & \cmark \\
BEHAVIOR-1K & CoRL'22 & \xmark & \xmark      & \cmark & \cmark & \cmark & \cmark            & \xmark & \cmark \\
ManiSkill-HUB & ICLR'25 & \xmark & \xmark      & \cmark & \cmark & \cmark & \cmark            &  \cmark &  \cmark \\
\midrule
RP2K & arXiv'20   & \cmark    & \xmark           & \xmark & \xmark & \xmark & \xmark            & \xmark      & \xmark      \\
SKU110K & CVPR'19    & \cmark    & \xmark           & \xmark & \xmark & \xmark & \xmark            & \xmark      & \xmark      \\
StandardSim & ICIAP'22 & \cmark    & \xmark     & \cmark & \xmark  & \xmark & \xmark            & \xmark      & \xmark      \\
IPA-3D1K & IROS'23  & \cmark    & \xmark & \cmark & \cmark & \xmark & \xmark            & \xmark      & \xmark      \\

FetchBot & arXiv'25   & \cmark    & \xmark           & \cmark & \xmark & \cmark & \xmark & \cmark      & \xmark       \\
\midrule
\RBM{}    &    & \cmark    & \cmark           & \cmark & \cmark & \cmark & \cmark            & \cmark      & \cmark    \\
\bottomrule
\end{tabular}}
\caption{Comparing proposed robotics retail benchmark with other benchmarks and datasets. 
A comprehensive review of related works can be found in Appendix~\ref{sec:related_work}.
}
\label{tab:review:bench}
\end{table*}

At the same time, retail environments pose challenges not captured by current benchmarks: operation in narrow, cluttered aisles, multi-level shelving, mixed fixtures (shelves, fridges), a large variety of visually similar products, and the need for reliable collision avoidance to protect inventory, infrastructure, and the robot itself.
As testing policies directly in these settings is difficult and expensive, high-fidelity simulation benchmarks that capture these properties are essential.

To address these gaps and advance research in robotic retail automation, we introduce \textbf{\RBM}---an open-source simulated retail benchmark suite designed to reflect the complexities of real-world retail tasks.

Our main contributions are as follows:
\begin{itemize}
    \item We introduce \textit{Store Plan Generator}, an open procedural pipeline for generating realistic and diverse store layouts and product arrangements, enabling scalable creation of retail environments for training and evaluating robotic policies.
    \item We present \textit{Store Trajectories Sampler}, a pipeline that automatically collects trajectories for common retail tasks using motion planning, and we release a dataset of synthetic trajectories generated for the Fetch robot embodiment.
    \item We introduce \textit{Store Robotics Benchmark}, to the best of our knowledge the first open benchmark dedicated to evaluating robotic policies in retail environments.
    Using our benchmark, we demonstrate that current state-of-the-art ``generalist'' models, even after realistic adaptation, struggle to complete typical retail tasks.
\end{itemize}
\section{Store Plan Generator}

To emulate darkstore environments, our simulation scenes are designed as warehouse-like spaces containing shelving units and refrigerators arranged in various configurations.
To facilitate domain randomization, we apply diverse textures to walls, floors, and ceilings, and incorporate multiple fixture designs.
Product items are placed on the shelves in realistic, randomized positions.

\subsection{Fixture Arrangement}

\begin{figure*}[hbt!]
    \centering
    \begin{subfigure}[t]{0.3\textwidth}
        \centering
        \includegraphics[width=\textwidth]{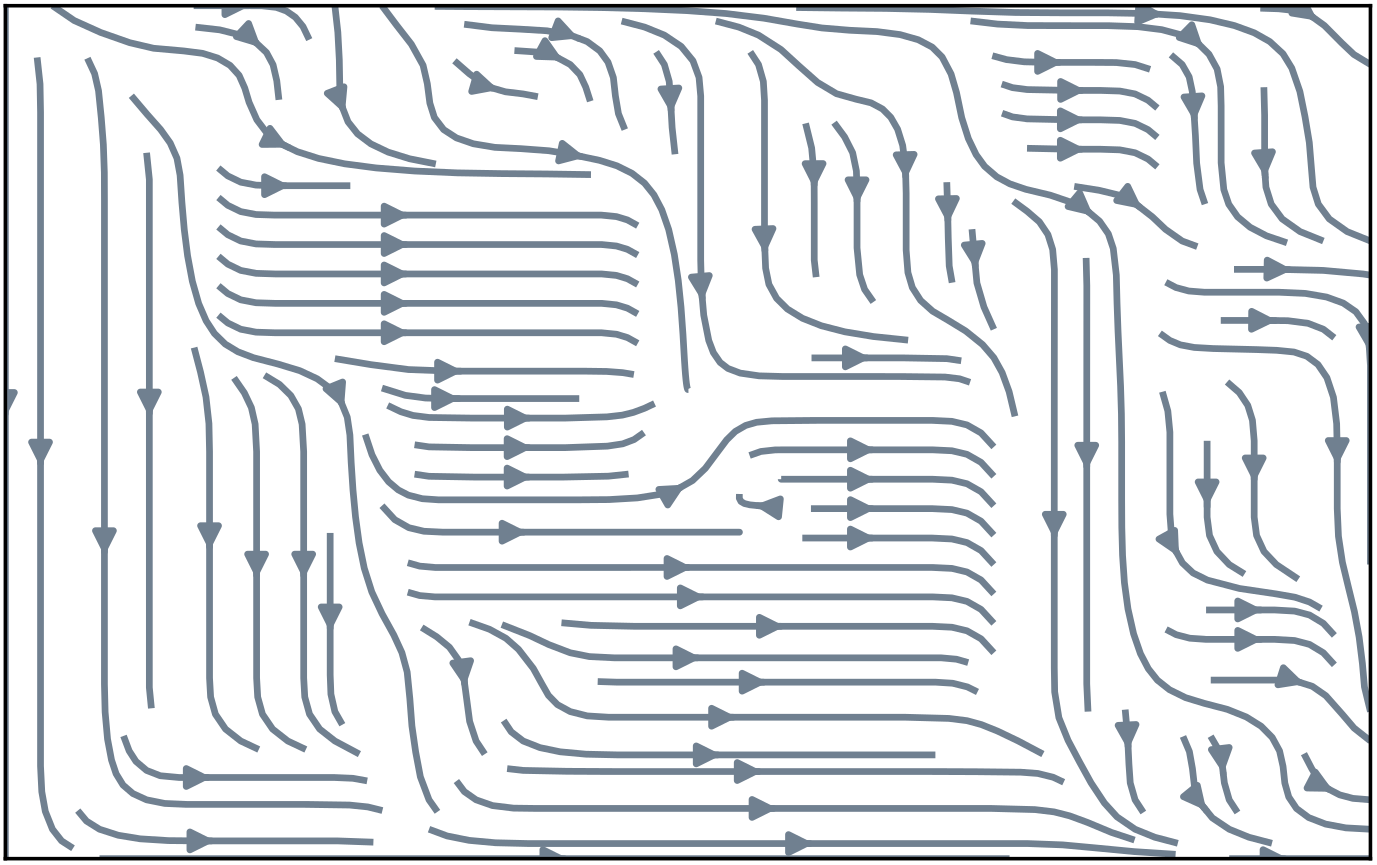}
        \caption{Sampled tensor field}
    \end{subfigure}
    \begin{subfigure}[t]{0.3\textwidth}
        \centering
        \includegraphics[width=\textwidth]{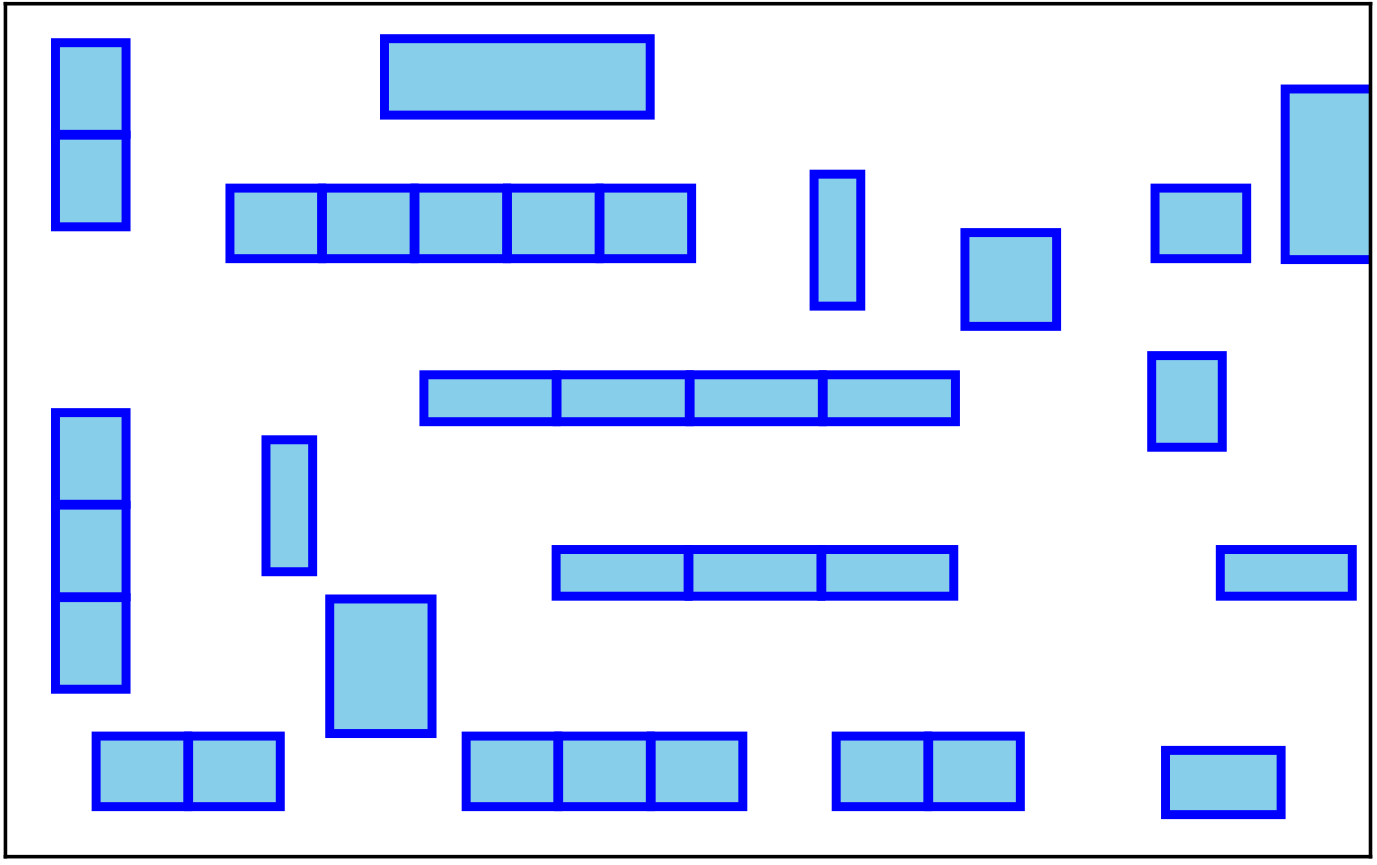}
        \caption{Resulted fixture layout}
    \end{subfigure}
    \begin{subfigure}[t]{0.3\textwidth}
        \centering
        \includegraphics[width=\textwidth]{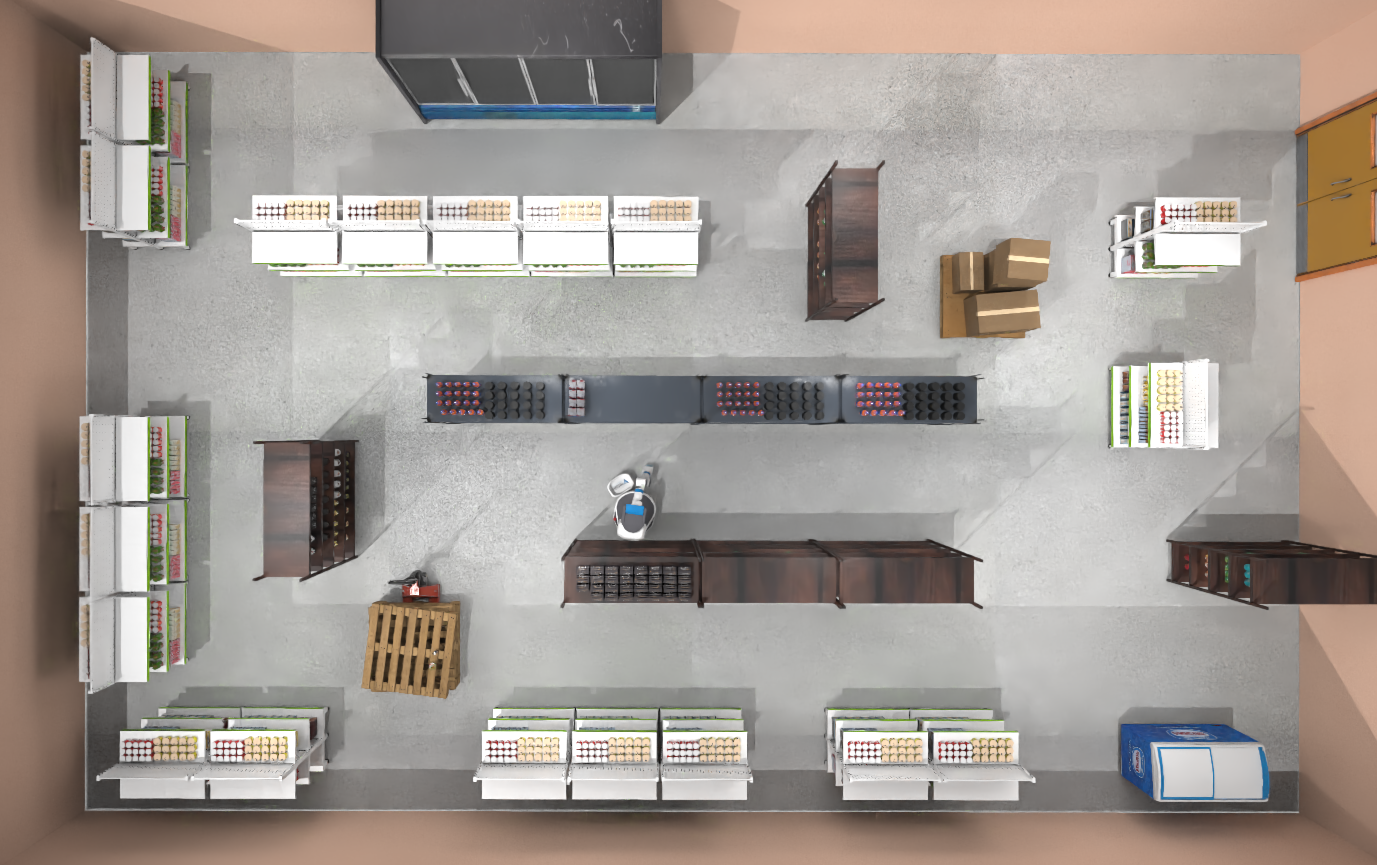}
        \caption{Generated store}
    \end{subfigure}
    \caption{Examples of generated store with fixtures arranged by our pipeline.}
    \label{fig:store-gen}
\end{figure*}

Our store fixture arrangement pipeline is inspired by procedural street modeling~\cite{chen2008interactive} and consists of three main stages.

In the first \textbf{fixture random placement} stage, we generate a rectangular store area populated with randomly placed fixtures such as pallets, boxes, and freezers.
Initial placement is performed using rejection sampling to ensure collision-free poses.

In the second stage, a \textbf{store tensor field}~\cite{chen2008interactive} is computed.
Given an $N \times M$ square-meter store floor and a list of initial fixtures, we compute the tensor field as follows:
\begin{enumerate}[label=(\arabic*)]
\item \textit{Polygon Construction:}
We extract polygons for the store floor and each already placed fixture.
Each polygon is defined as a closed sequence of vertices $\{ \boldsymbol{p}_i\}_{i=1}^{P} $, where each point $\boldsymbol{p}_i$ is connected to $\boldsymbol{p}_{i+1}$, and for $i = P$, we define $\boldsymbol{p}_{P+1} \coloneqq \boldsymbol{p}_1$.
We also ensure that for all edge vectors $\boldsymbol{u}_i \coloneqq \boldsymbol{p}_i - \boldsymbol{p}_{i+1}$, the condition $||\boldsymbol{u}_i|| \leq D$ holds, where $D$ denotes the maximum allowed edge length.

\item \textit{Basis Tensor Field Computation:}

For each point $\boldsymbol{p}_i$ on the polygon, we compute a basis tensor:
$T(\boldsymbol{p}) = l \, \begin{pmatrix}
    \cos{2 \, \theta} & \sin{2 \, \theta} \\
    \sin{2 \, \theta} & -\cos{2 \, \theta}
\end{pmatrix}$, where $l=||\boldsymbol{u}_i||$, $\theta = \arctan{\frac{u_{ix}}{u_{iy}}}$.
This results in a set of basis tensors $\{T_j(\boldsymbol{p})\}$.

\item \textit{Tensor Field Aggregation:}
The final tensor field over the store layout is computed as a weighted sum of the basis tensors:
$T(\boldsymbol{p}) = \sum_j e^{-d \, ||\boldsymbol{p} - \boldsymbol{p}_j||} \, T_j(\boldsymbol{p})$, where the weights decay exponentially with the distance from $\boldsymbol{p}$ to the basis point $\boldsymbol{p}_j$, and the parameter $d > 0$ controls the rate of decay.

\end{enumerate}

In the final \textbf{shelving unit arrangement} stage, shelves are placed according to the tensor field, ensuring alignment with local directions and maintaining a collision-free, navigable layout.
Placement is performed in two passes:
\begin{enumerate*}[label=(\arabic*)]
\item \textit{Horizontal pass:} Shelving units are placed row by row at positions where the tensor field indicates a horizontal orientation.
\item \textit{Vertical pass:} Shelving units are placed column by column where the field indicates a vertical orientation.
\end{enumerate*}
At each step, the local direction is interpolated from the precomputed tensor grid, and placements are accepted only if they are collision-free and maintain the required passage width.
Probabilistic skipping adds layout variability.

The proposed pipeline generates a structured and realistic store layout guided by the underlying sampled tensor field (see Figure~\ref{fig:store-gen}).


\subsection{Product Arrangement}

\begin{figure*}[tbh!]
    \centering

    \centering
    \begin{subfigure}[t]{0.23\textwidth}
        \centering
        \includegraphics[width=1\textwidth]{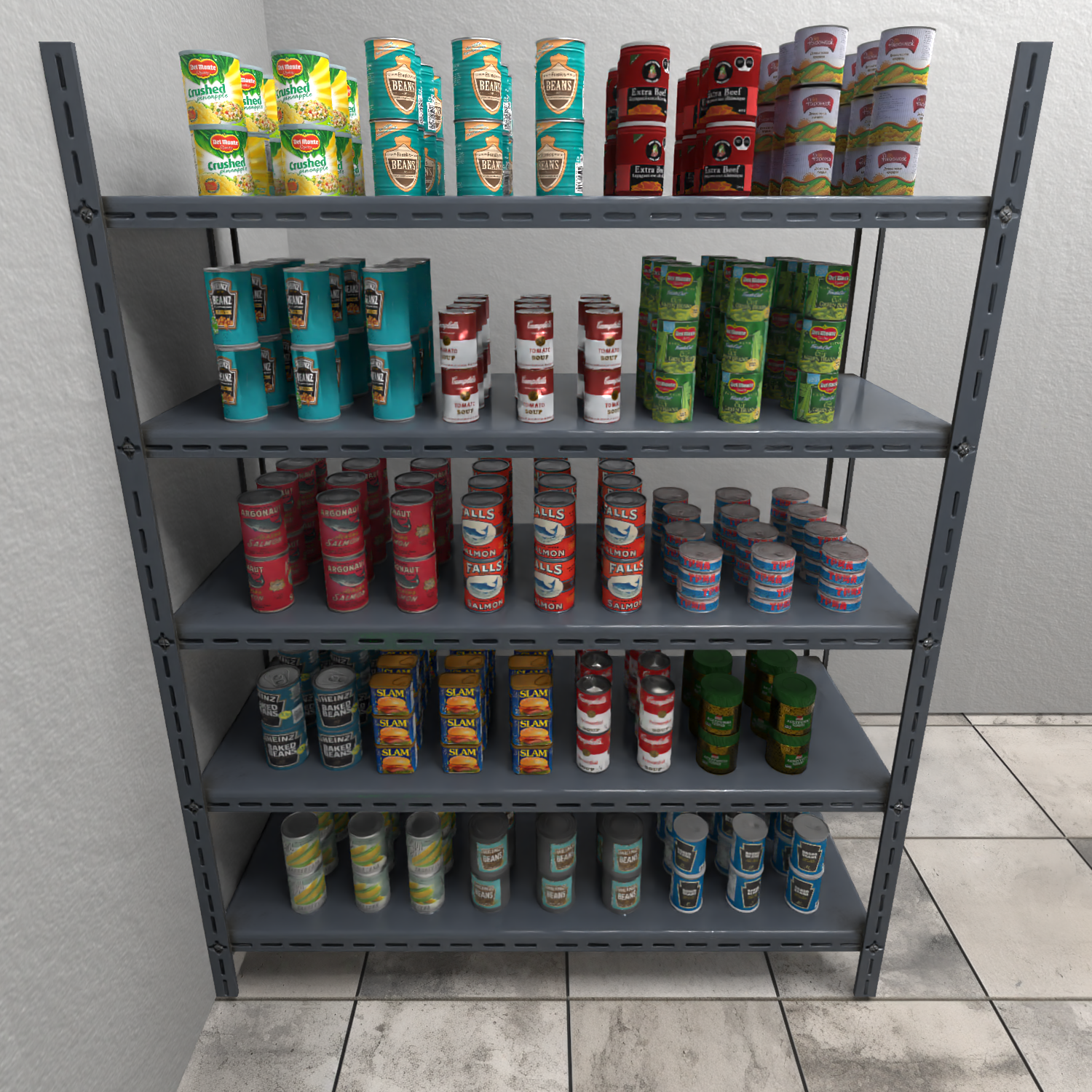}
        \caption{1$^{st}$ day}
        \label{fig:product-arrangement:day1}
    \end{subfigure}
    \begin{subfigure}[t]{0.23\textwidth}
        \centering
        \includegraphics[width=1\textwidth]{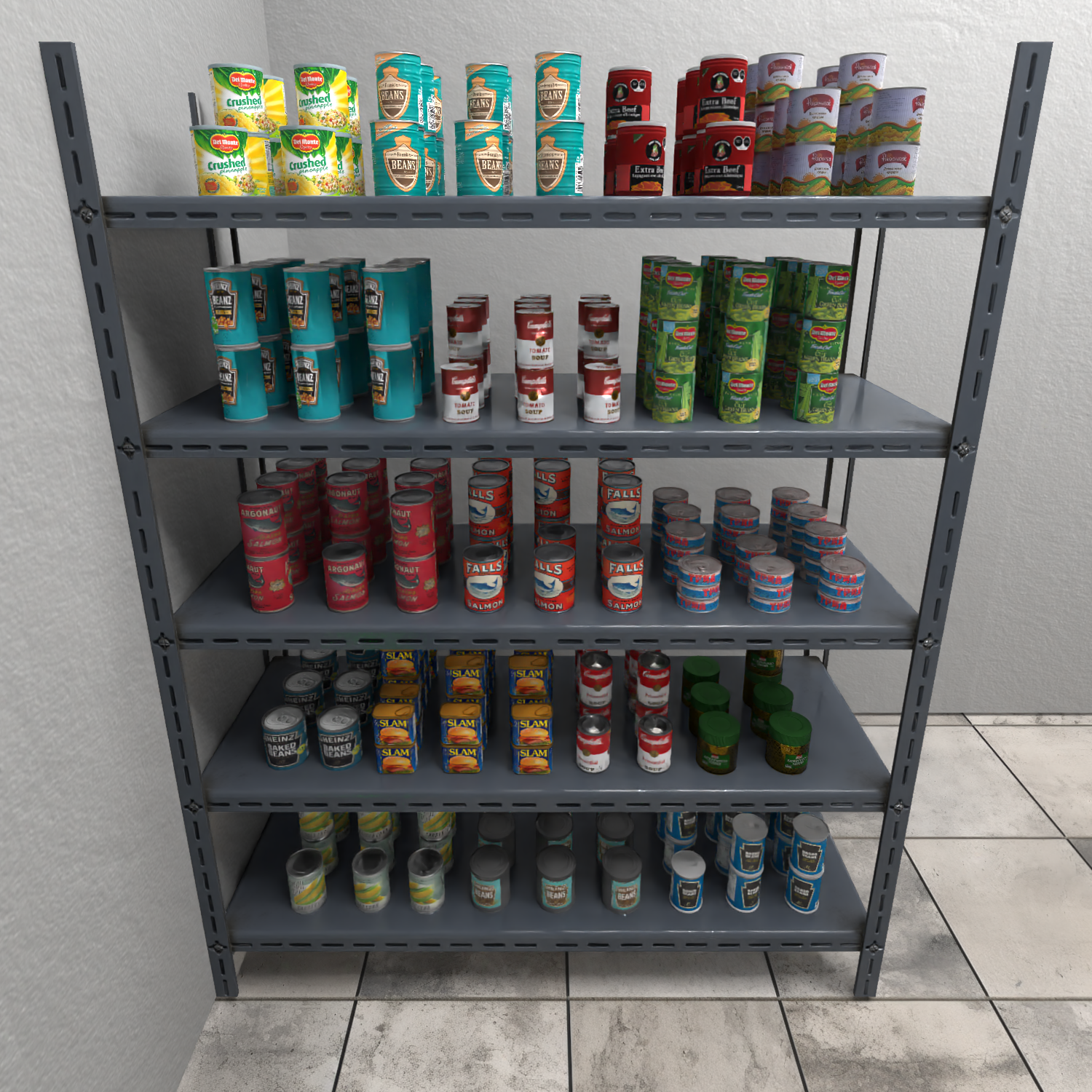}
        \caption{2$^{nd}$ day}
    \end{subfigure}
    \begin{subfigure}[t]{0.23\textwidth}
        \centering
        \includegraphics[width=1\textwidth]{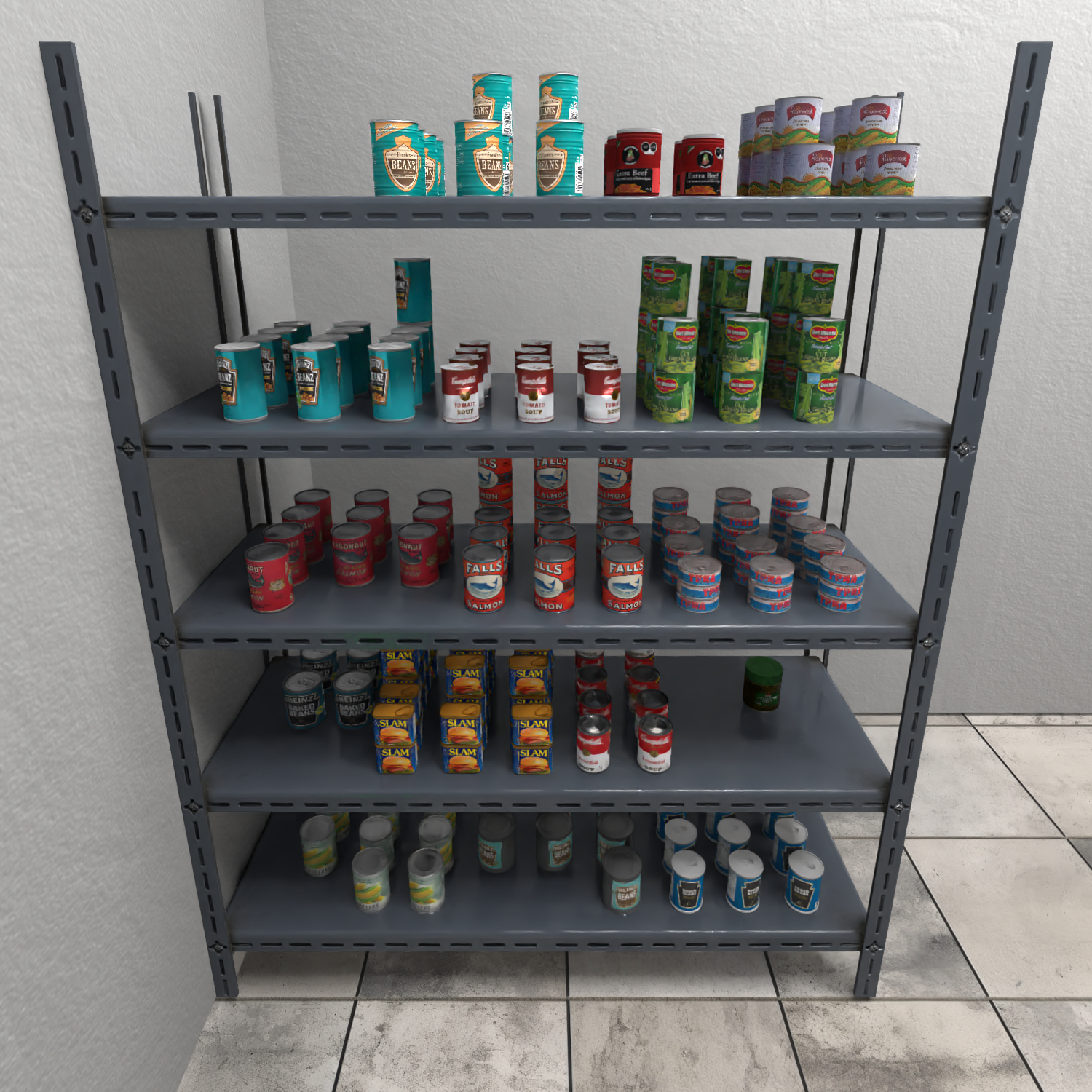}
        \caption{4$^{th}$ day}
    \end{subfigure}
    \begin{subfigure}[t]{0.23\textwidth}
        \centering
        \includegraphics[width=1\textwidth]{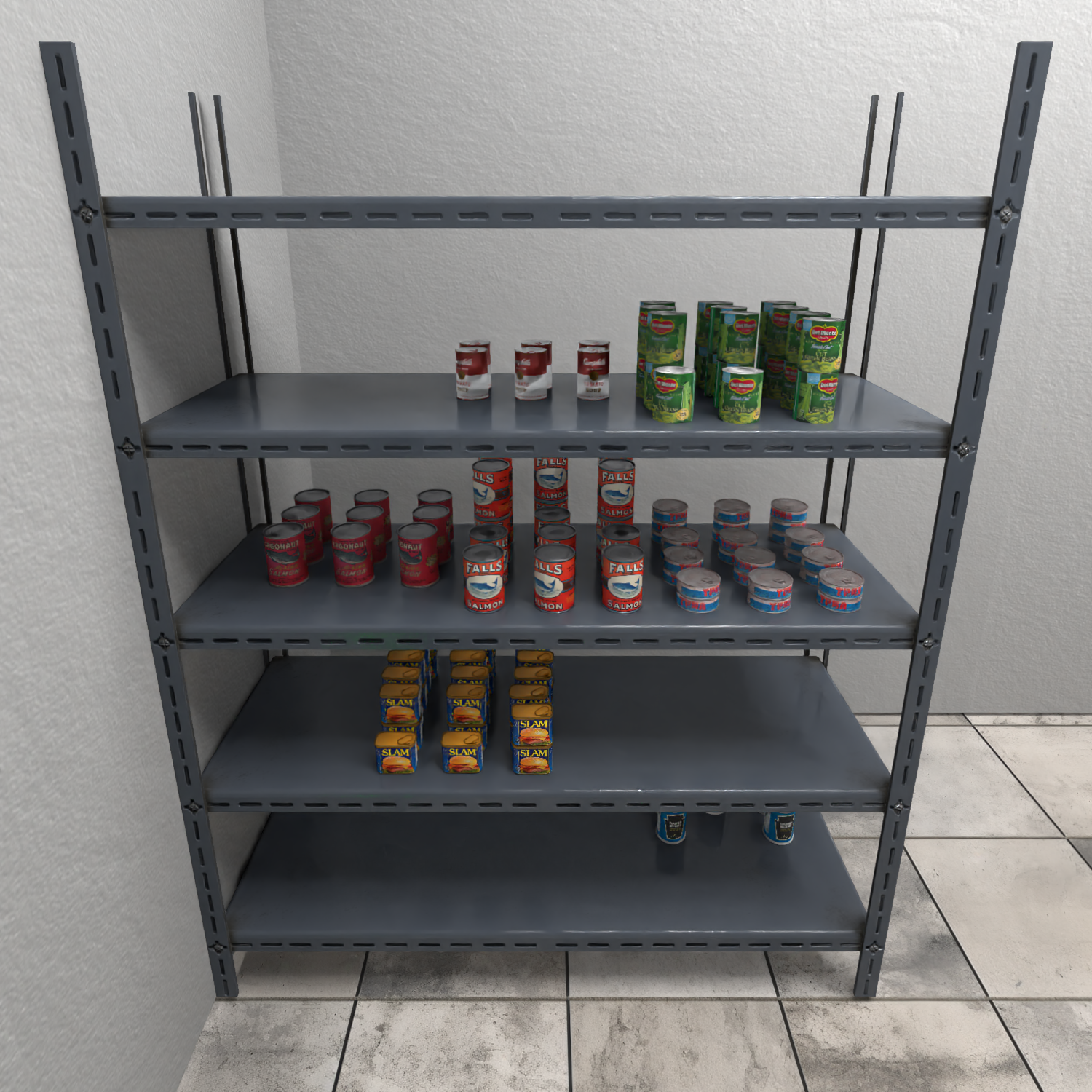}
        \caption{8$^{th}$ day}
    \end{subfigure}
    \caption{Example of product arrangement and shelf depletion over time produced by our simulator.}
    \label{fig:product-arrangement}
\end{figure*}

To arrange products on shelves, we leverage the \texttt{scene\_synthesizer}~\cite{eppner2025scene} package, which automatically detects shelf surfaces suitable for item placement.
Building on this framework, we implement a custom product placement module that positions items on a regular grid while introducing small pose perturbations to simulate natural variability.
The module also supports vertical stacking of items, as commonly observed in real stores.
Additionally, it can leave empty spaces at the front of shelves using a Poisson process, simulating natural product depletion over time.
Examples of product arrangements generated by our module are shown in Figure~\ref{fig:product-arrangement}.

\begin{figure*}[tbh!]
    \centering

    \centering
    \begin{subfigure}[t]{0.23\textwidth}
        \centering
        \includegraphics[width=1\textwidth]{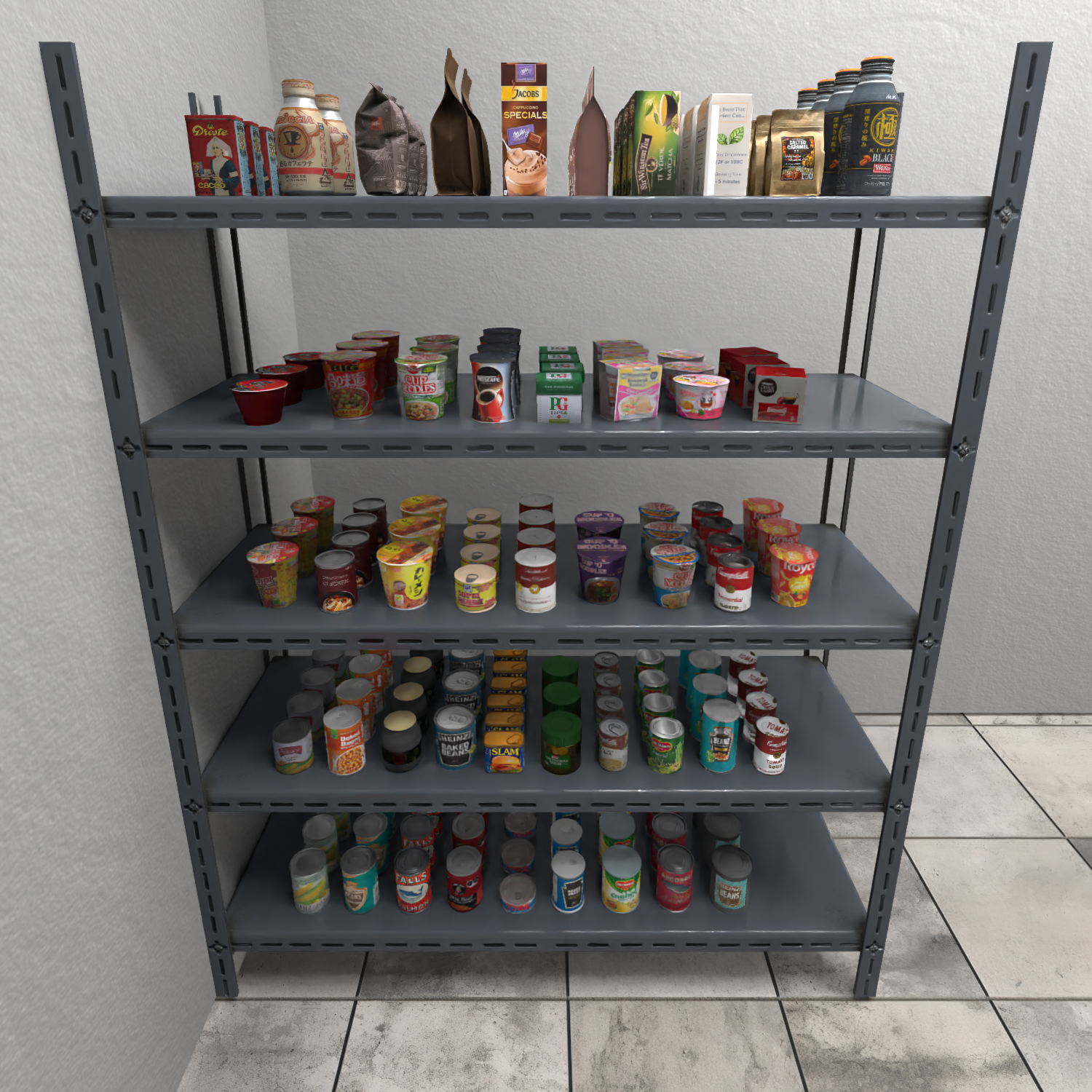}
    \end{subfigure}
    \begin{subfigure}[t]{0.23\textwidth}
        \centering
        \includegraphics[width=1\textwidth]{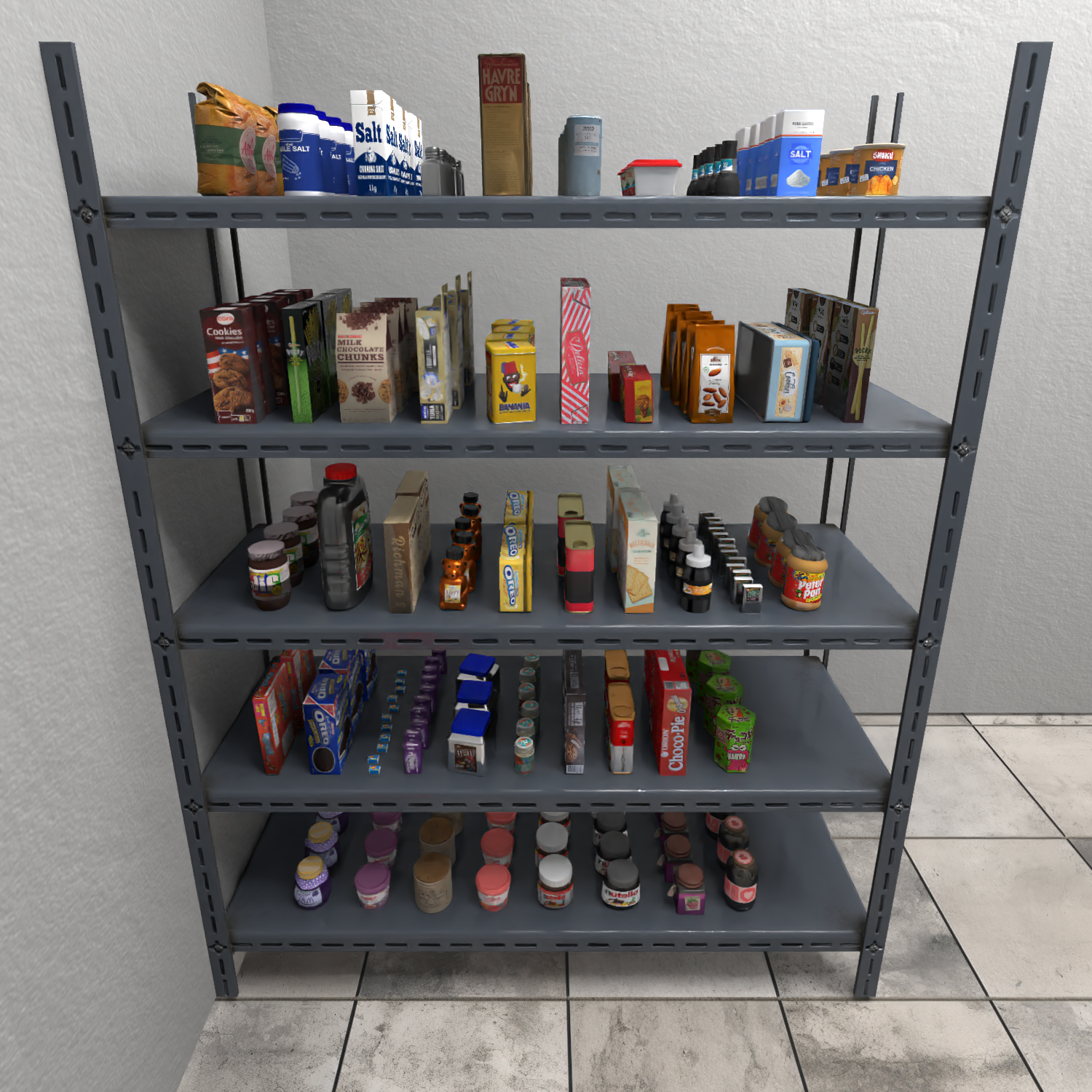}
    \end{subfigure}
    \begin{subfigure}[t]{0.23\textwidth}
        \centering
        \includegraphics[width=1\textwidth]{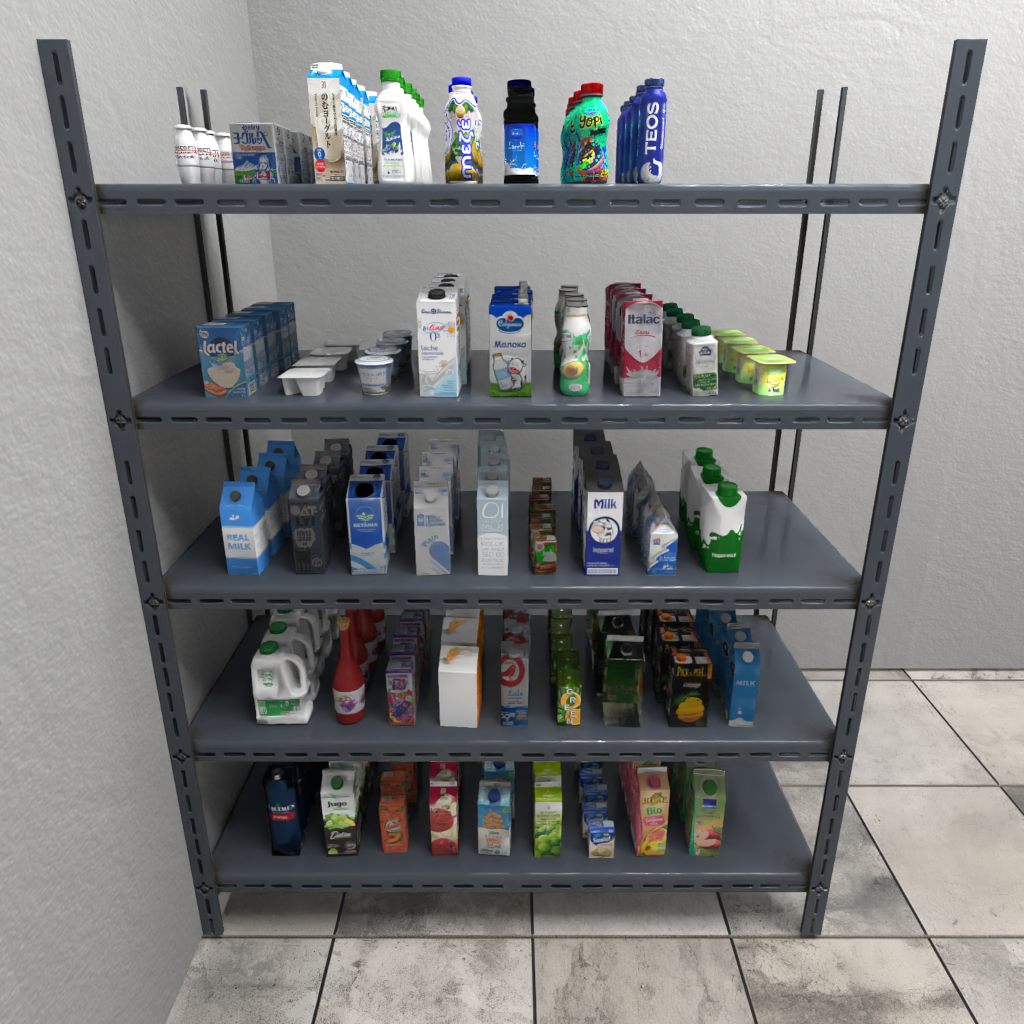}
    \end{subfigure}
    \begin{subfigure}[t]{0.23\textwidth}
        \centering
        \includegraphics[width=1\textwidth]{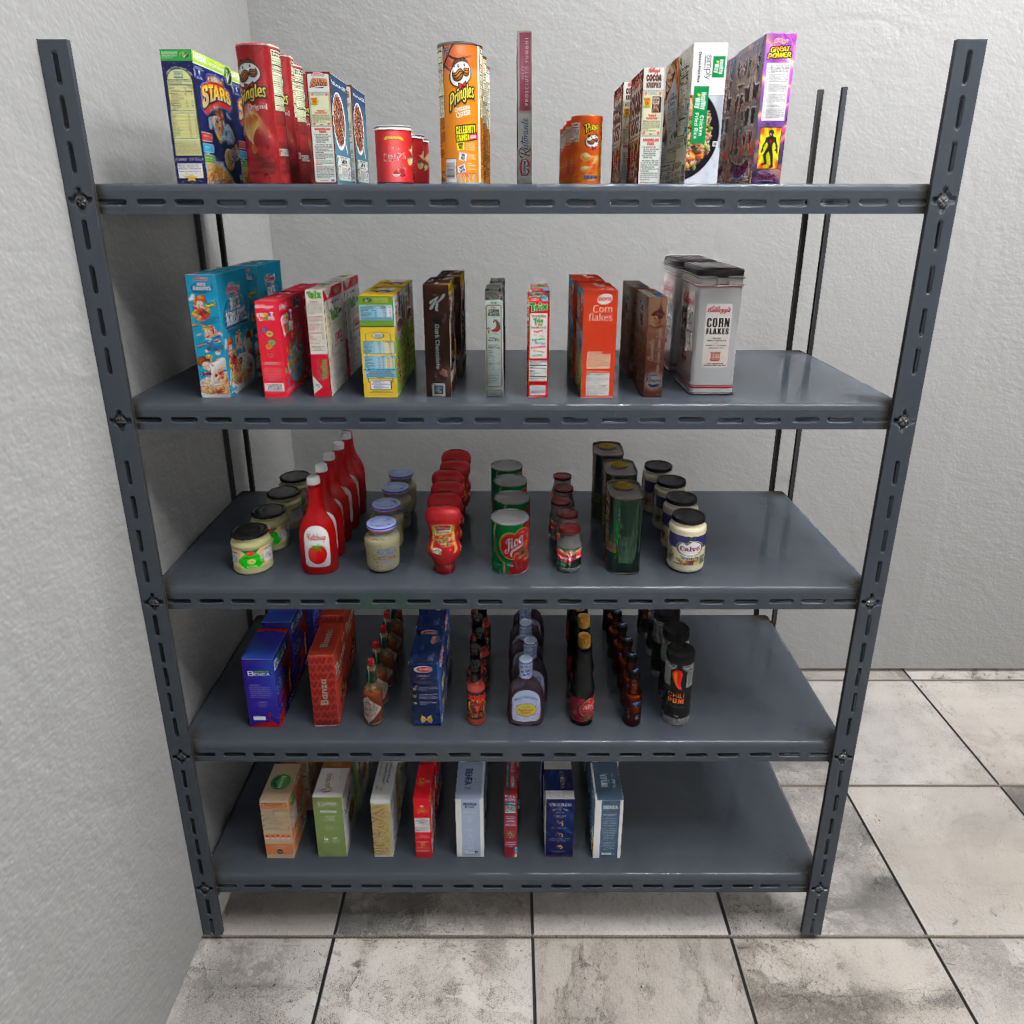}
    \end{subfigure}
    \caption{Examples of collected product assets.}
    \label{fig:collected-assets}
\end{figure*}

\vspace{-1em}

\subsection{Assets and Textures}
\label{sec:assets}

\begin{figure*}[hbt!]
    \centering
        \includegraphics[width=0.23\linewidth]{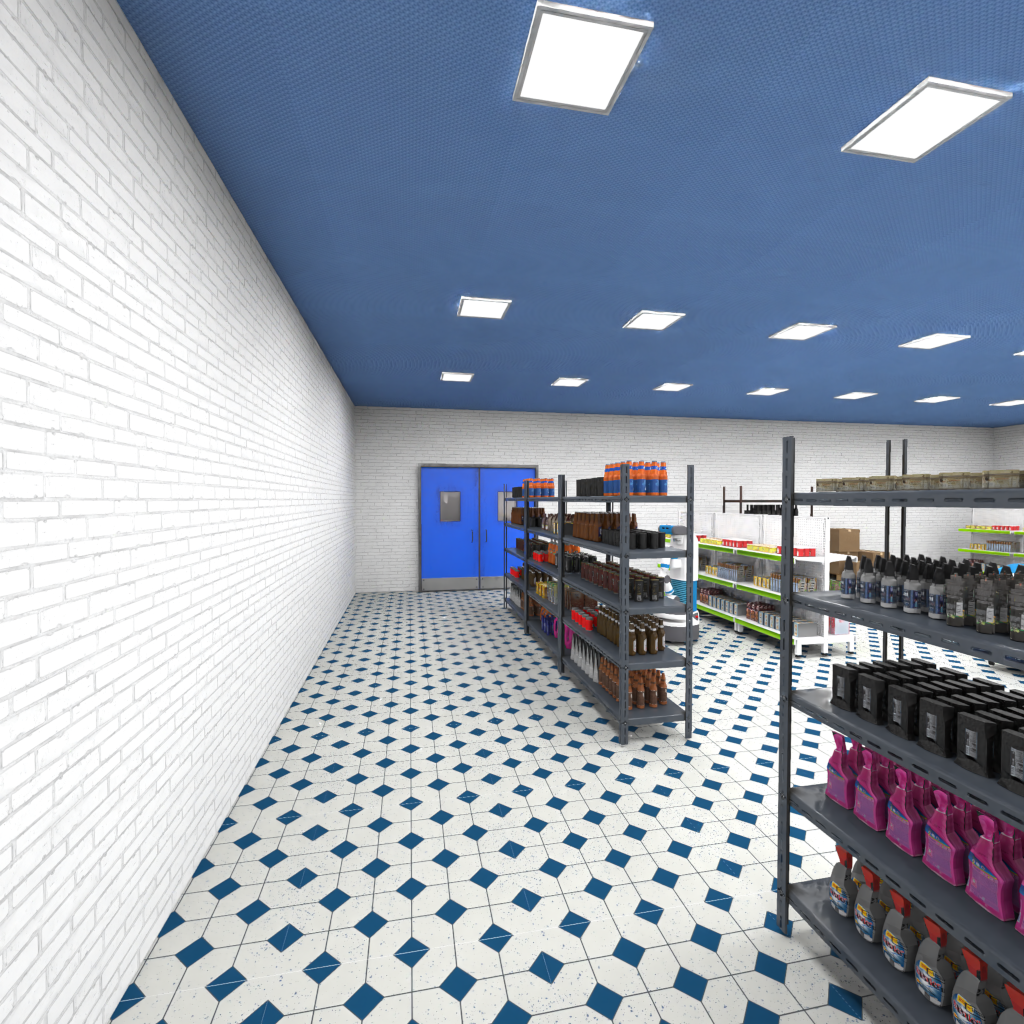}
        \includegraphics[width=0.23\linewidth]{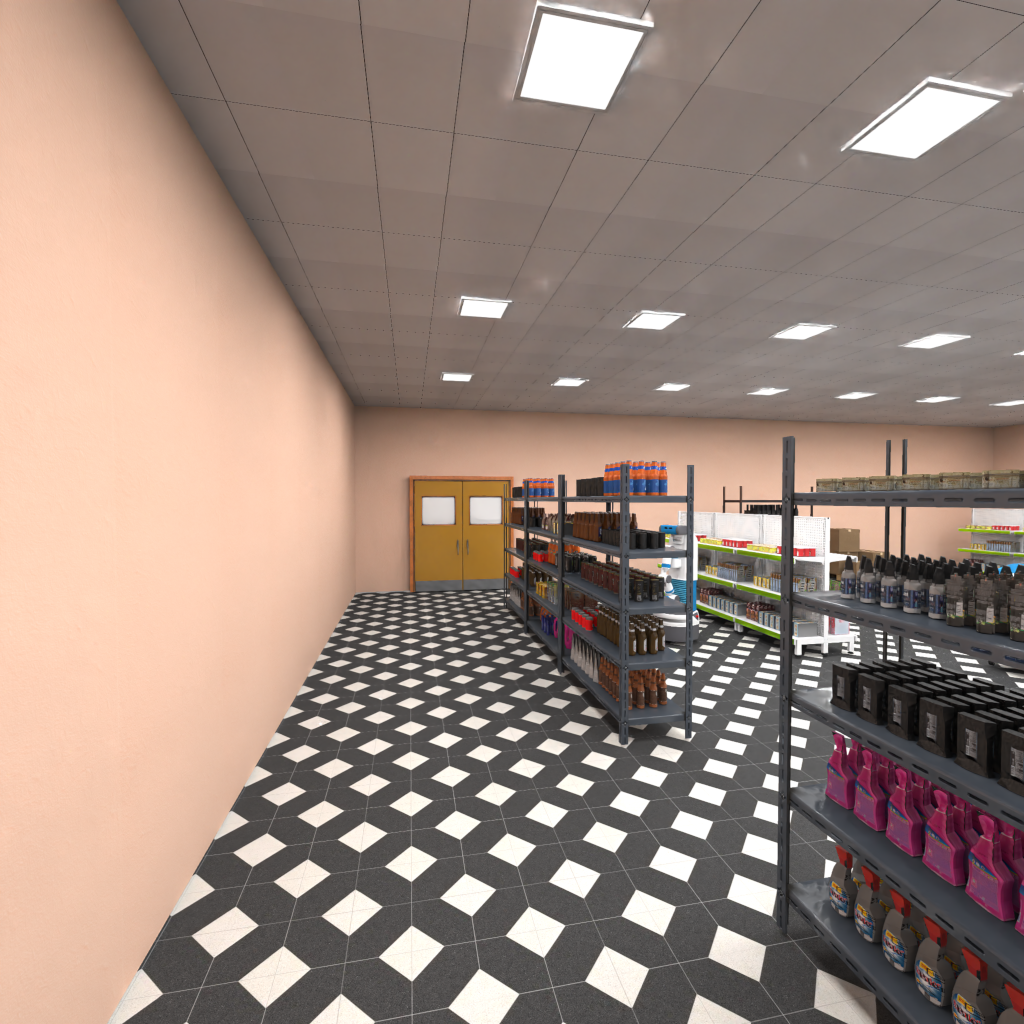}
        \includegraphics[width=0.23\linewidth]{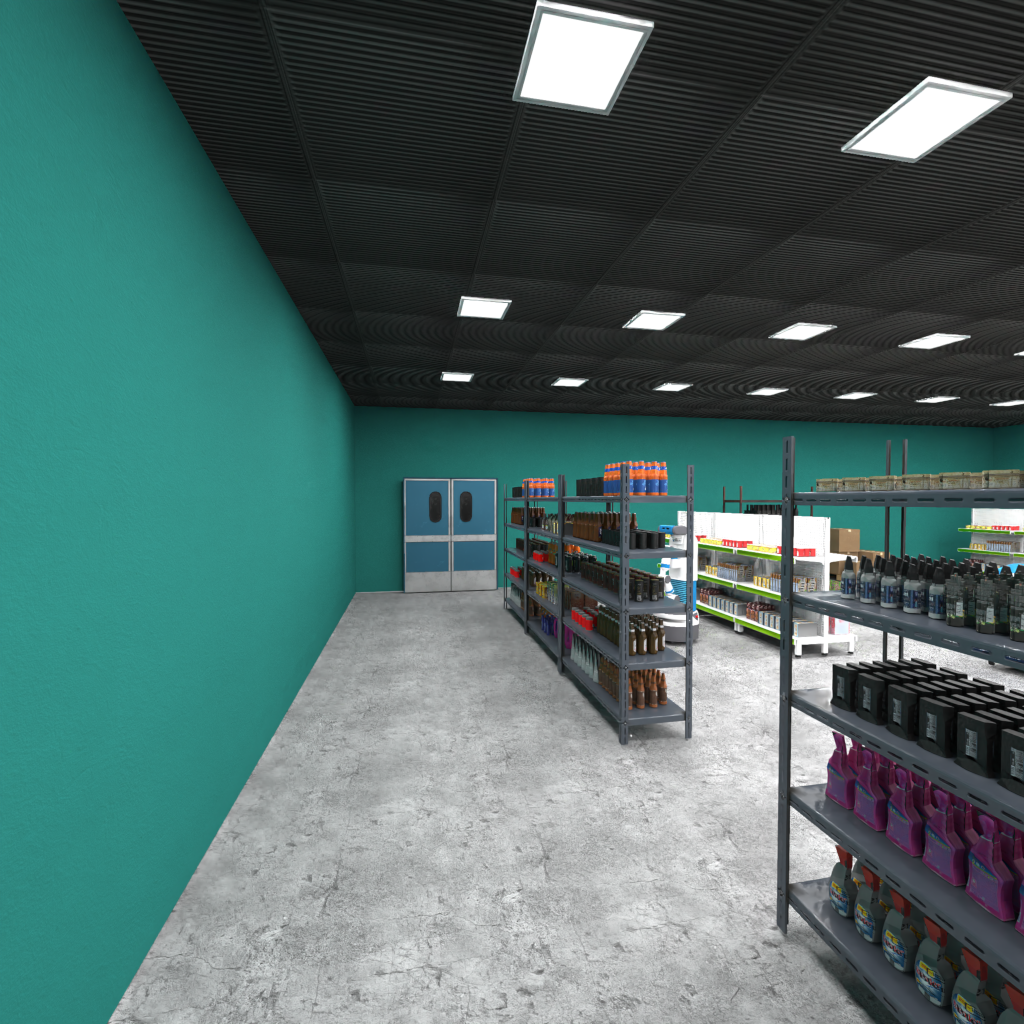}
        \includegraphics[width=0.23\linewidth]{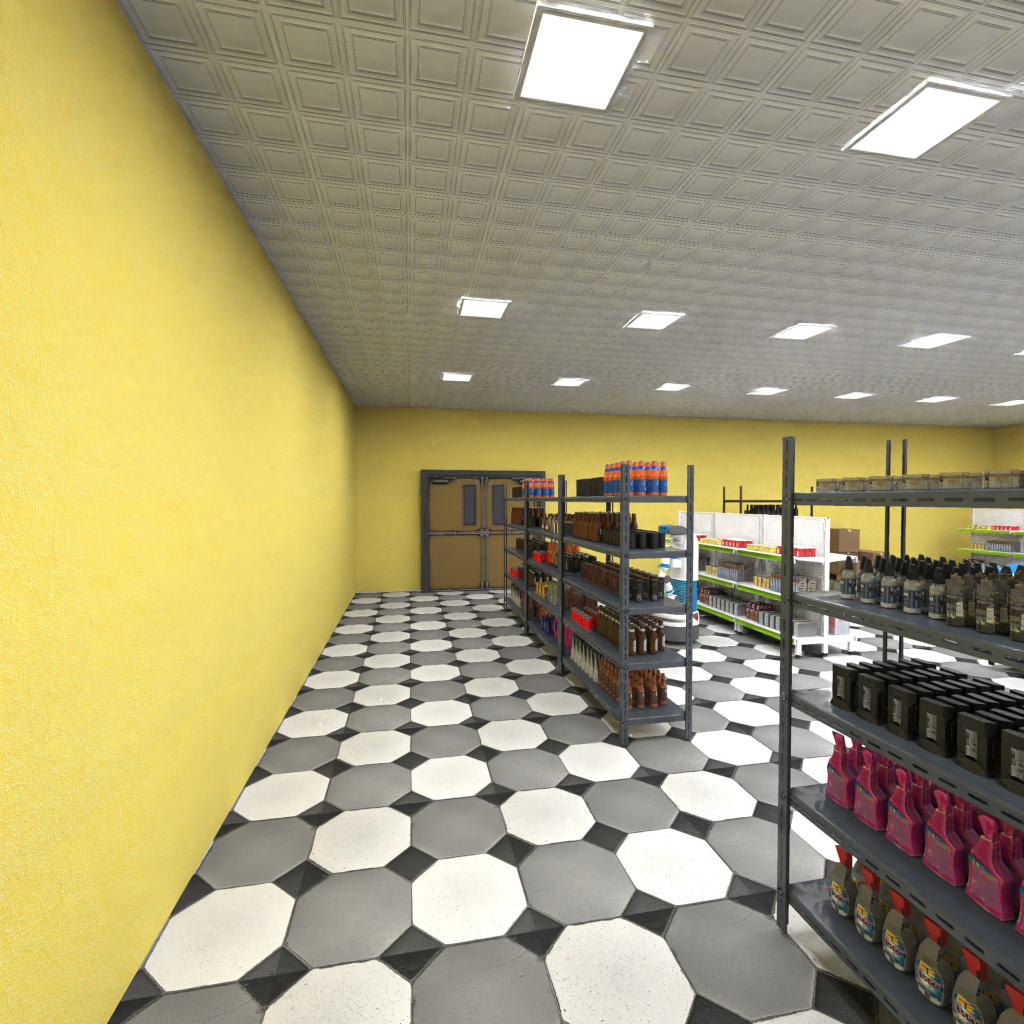}
    \caption{Examples of ceiling, wall, and floor textures used in our store generation pipeline, illustrating just a subset of possible variations.}
    \label{fig:layouts}
\end{figure*}

To represent 3D models of shelving units, refrigerators, and individual product items within our simulation environment, we utilize assets sourced from SketchFab\footnote{\url{https://sketchfab.com}}.
In total, we collected three shelving unit models, two refrigerator models, and 370 product assets across 21 categories (see Figure~\ref{fig:collected-assets}).
Additionally, we gathered 26 floor, 17 wall, and 15 ceiling textures to support visual diversity in the generated store environments (see Figure~\ref{fig:layouts}).
All assets and textures were verified to be licensed for unrestricted research use.

\todo{All collected and optimized assets are publicly available on \url{https://huggingface.co/datasets/emb-ai/RoboBenchMart_assets}.}


The collected 3D assets lacked consistent scale and orientation.
We manually standardized their orientation and adjusted each model's scale using reference dimensions obtained from online retail catalogs to ensure realistic proportions.

\begin{figure*}[tbh!]
    \centering
    \begin{minipage}[b]{0.62\textwidth}
        \centering
        \includegraphics[width=0.9\linewidth]{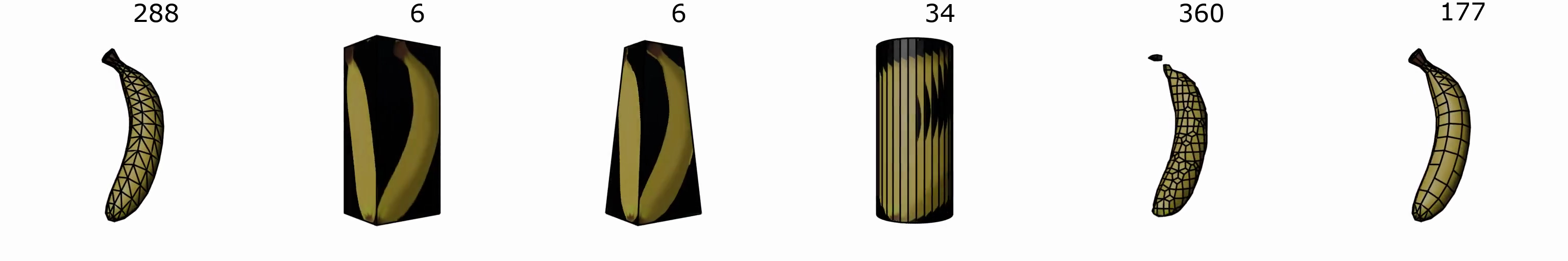} 
        \includegraphics[width=0.9\linewidth]{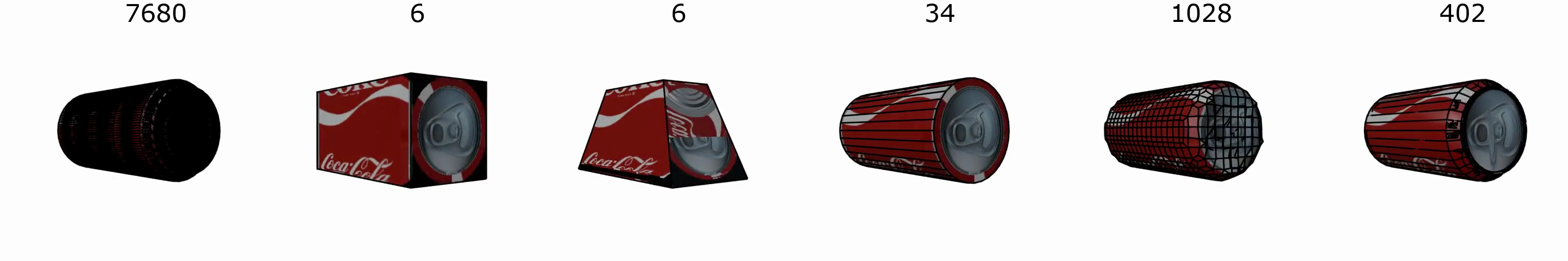}
        \includegraphics[width=0.9\linewidth]{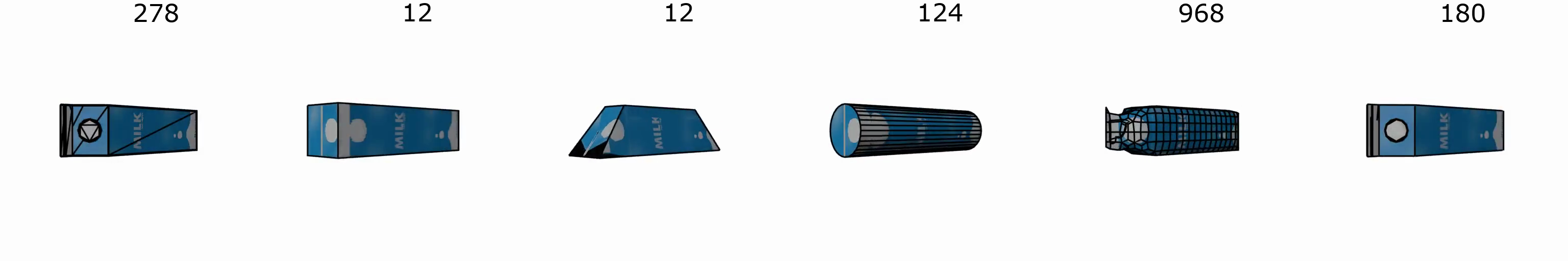}
        \captionof{figure}{Example of different geometry approximations for assets (original on the left). Numbers above indicate face count for each mesh.\newline\newline}
        \label{fig:assets:meshes}
    \end{minipage}
    \hspace{1em}
    \begin{minipage}[b]{0.34\textwidth}
        \centering
        \includegraphics[width=0.85\linewidth]{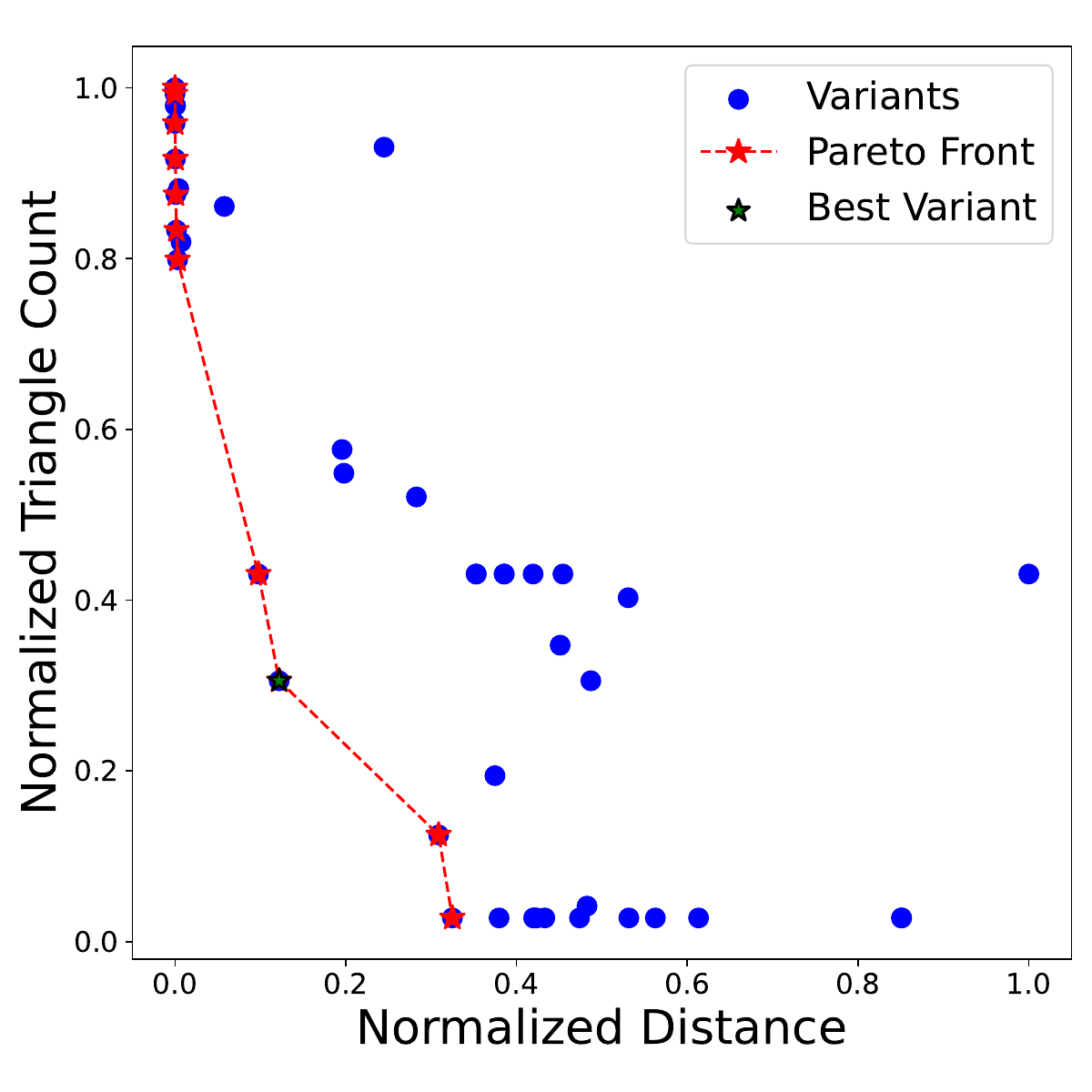}
        \captionof{figure}{Variety of generated simplified meshes. Distance and Triangle Count are given in relative units (w.r.t max distance and initial triangle count).}
        \label{fig:assets:pareto}
    \end{minipage}
\end{figure*}

Furthermore, many original product meshes were unoptimized and contained excessive triangle counts, which significantly slowed rendering when scenes included hundreds of objects. 
To improve performance, we developed an automatic mesh simplification pipeline.
Given the complexity of mesh decimation, an open research problem, we applied brute-force optimization across several heuristic methods, including QuadriFlow, Marching Cubes, and shape-specific approximations (e.g., cylinders, boxes), using the Blender Python API (see Figure~\ref{fig:assets:meshes}).
From the Pareto-optimal set of remeshed outputs, we selected the version minimizing the total $L_1$ relative drop in geometry quality (measured by Chamfer distance) and maximizing triangle reduction (see Figure~\ref{fig:assets:pareto}).

\vspace{-2em}

\section{Store Trajectories Sampler}

\begin{figure*}[tbh!]
    \centering
    \begin{subfigure}[t]{1\textwidth}
        \centering
        \includegraphics[width=0.19\textwidth]{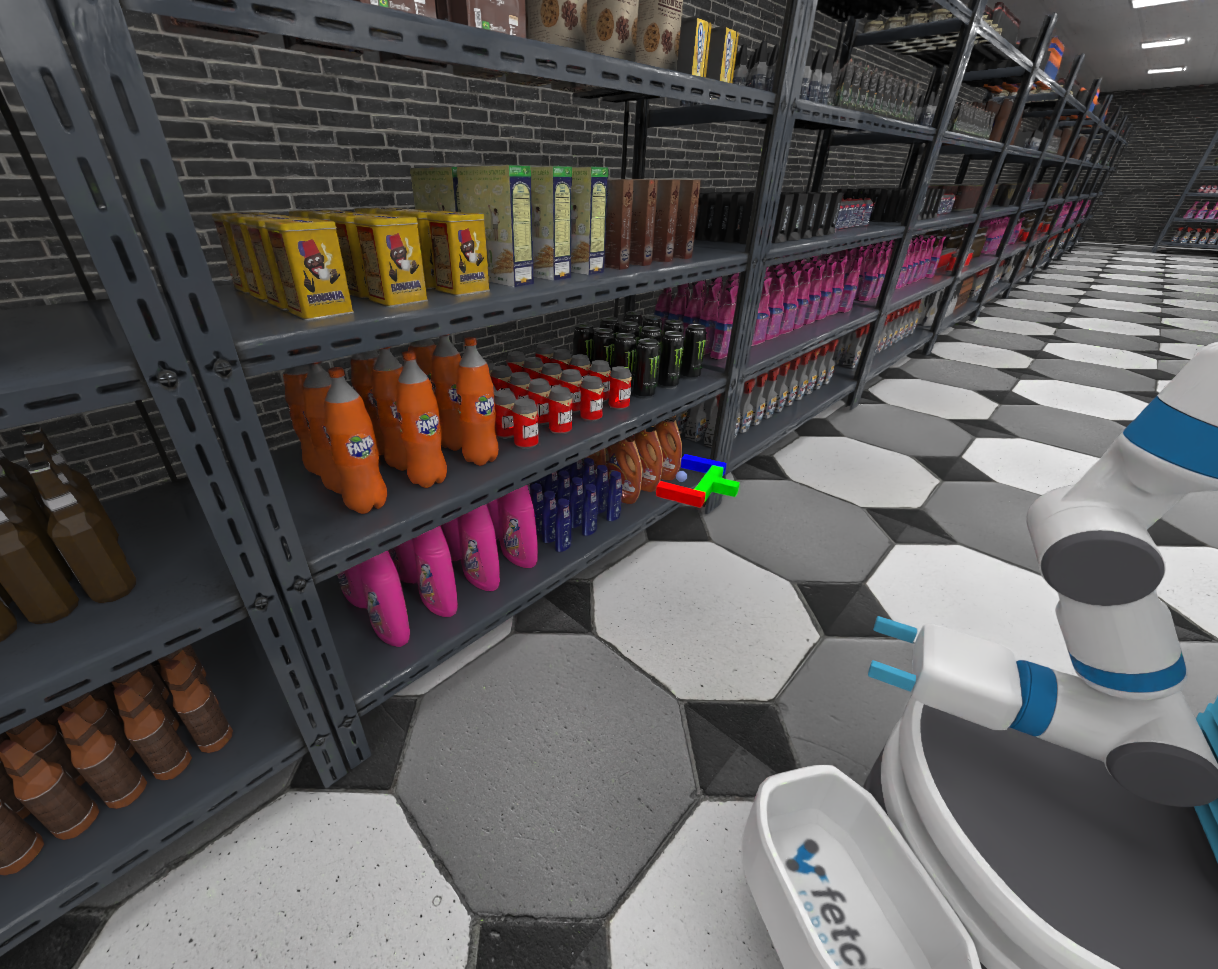}
        \includegraphics[width=0.19\textwidth]{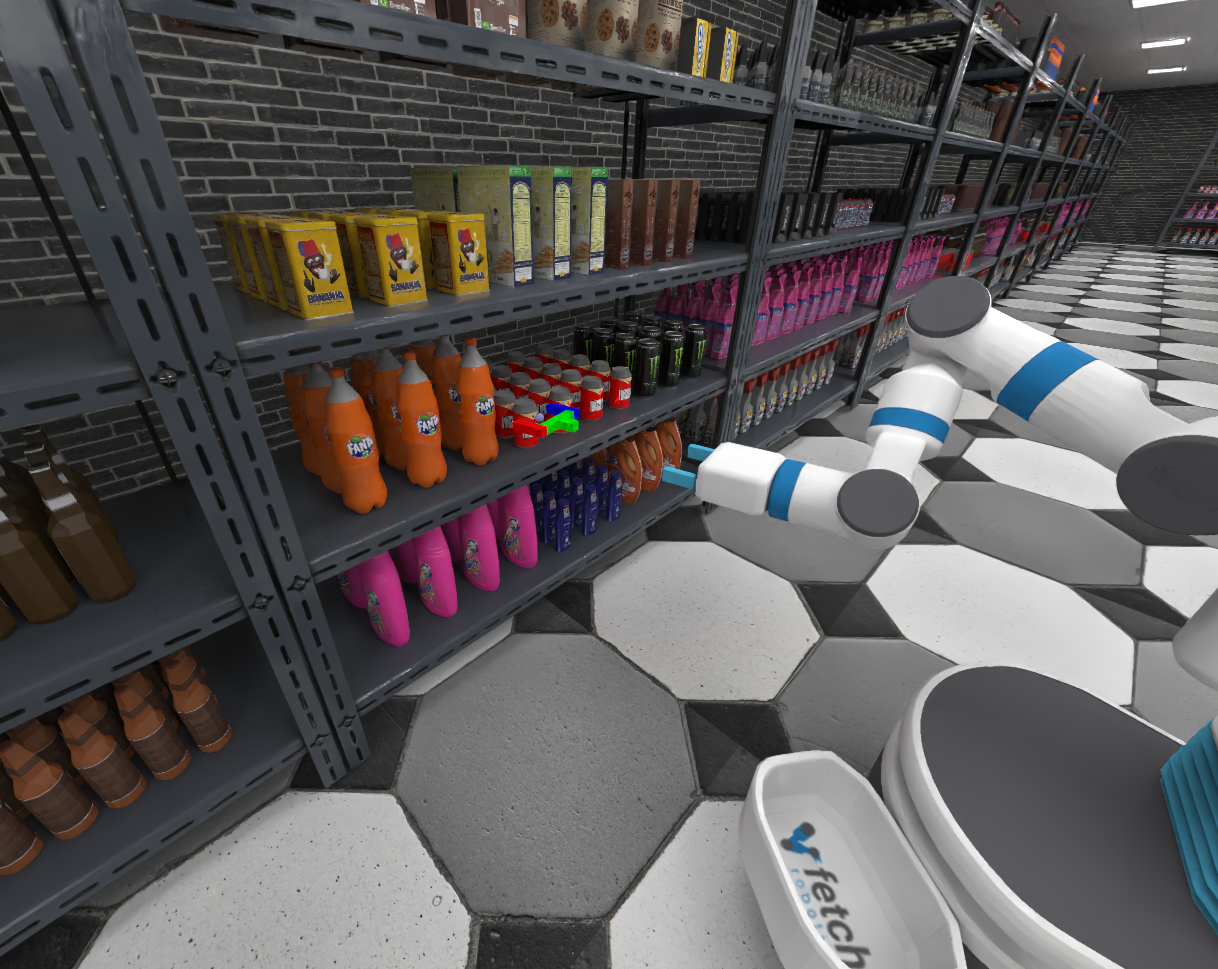}
        \includegraphics[width=0.19\linewidth]{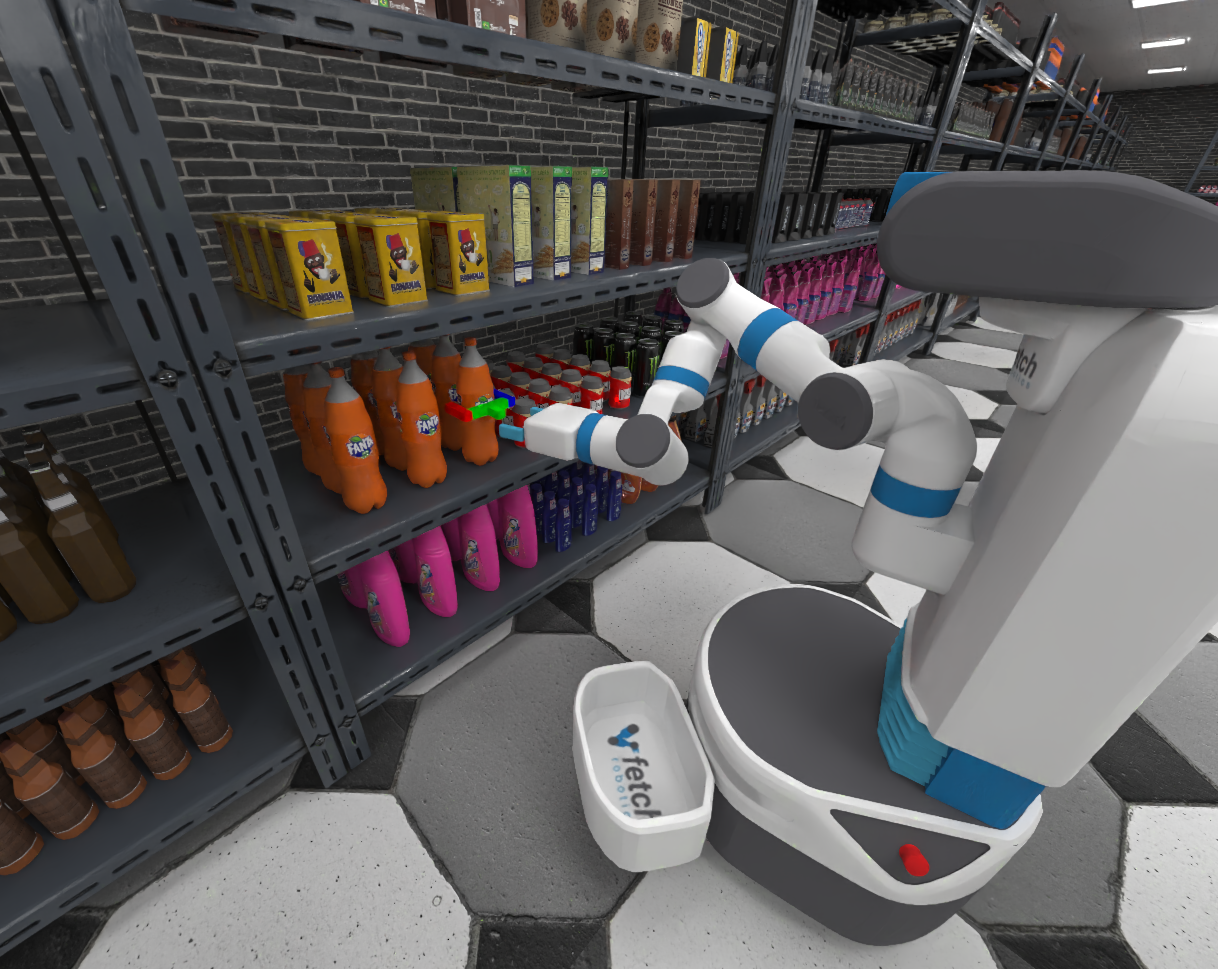}
        \includegraphics[width=0.19\linewidth]{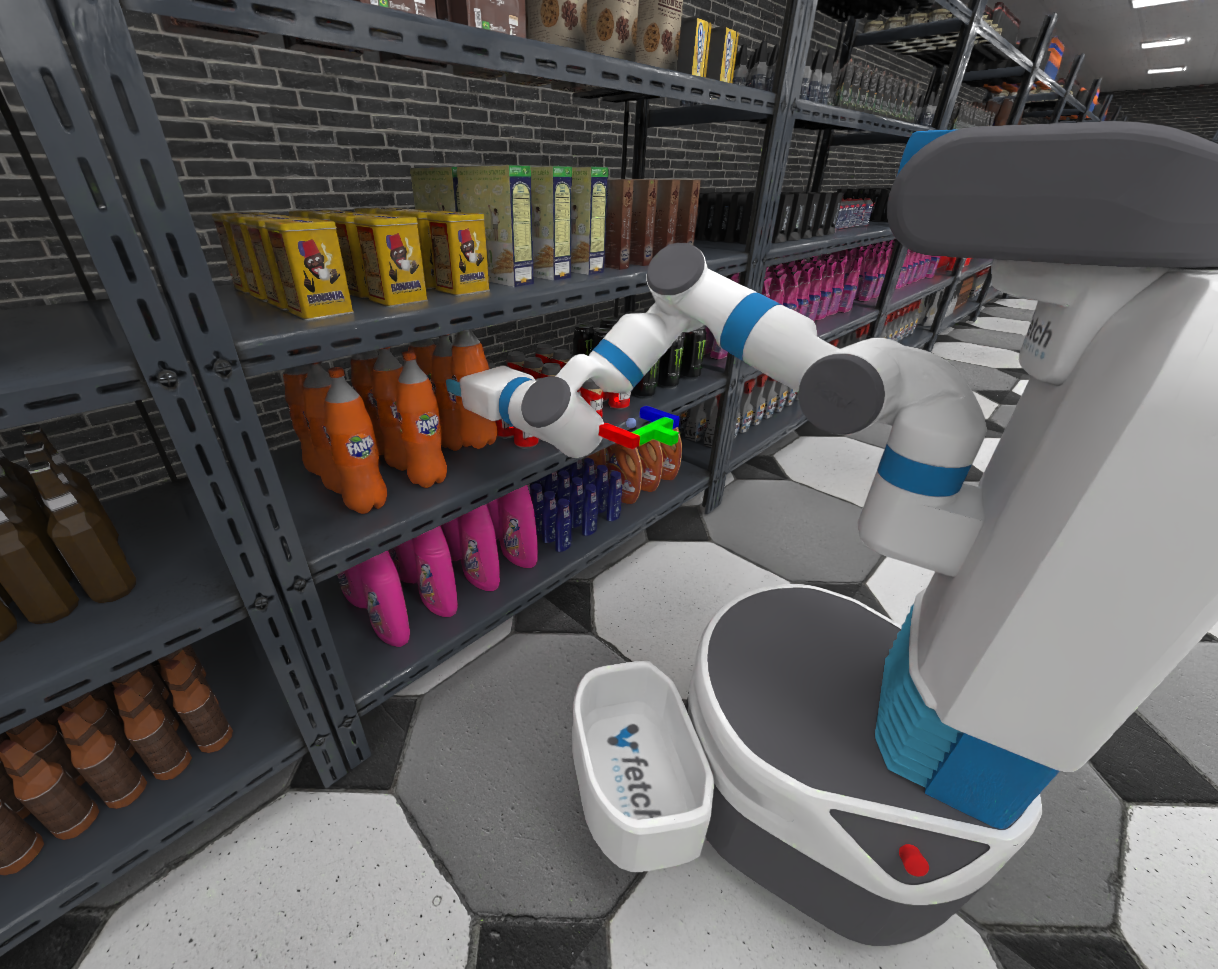}
        \includegraphics[width=0.19\linewidth]{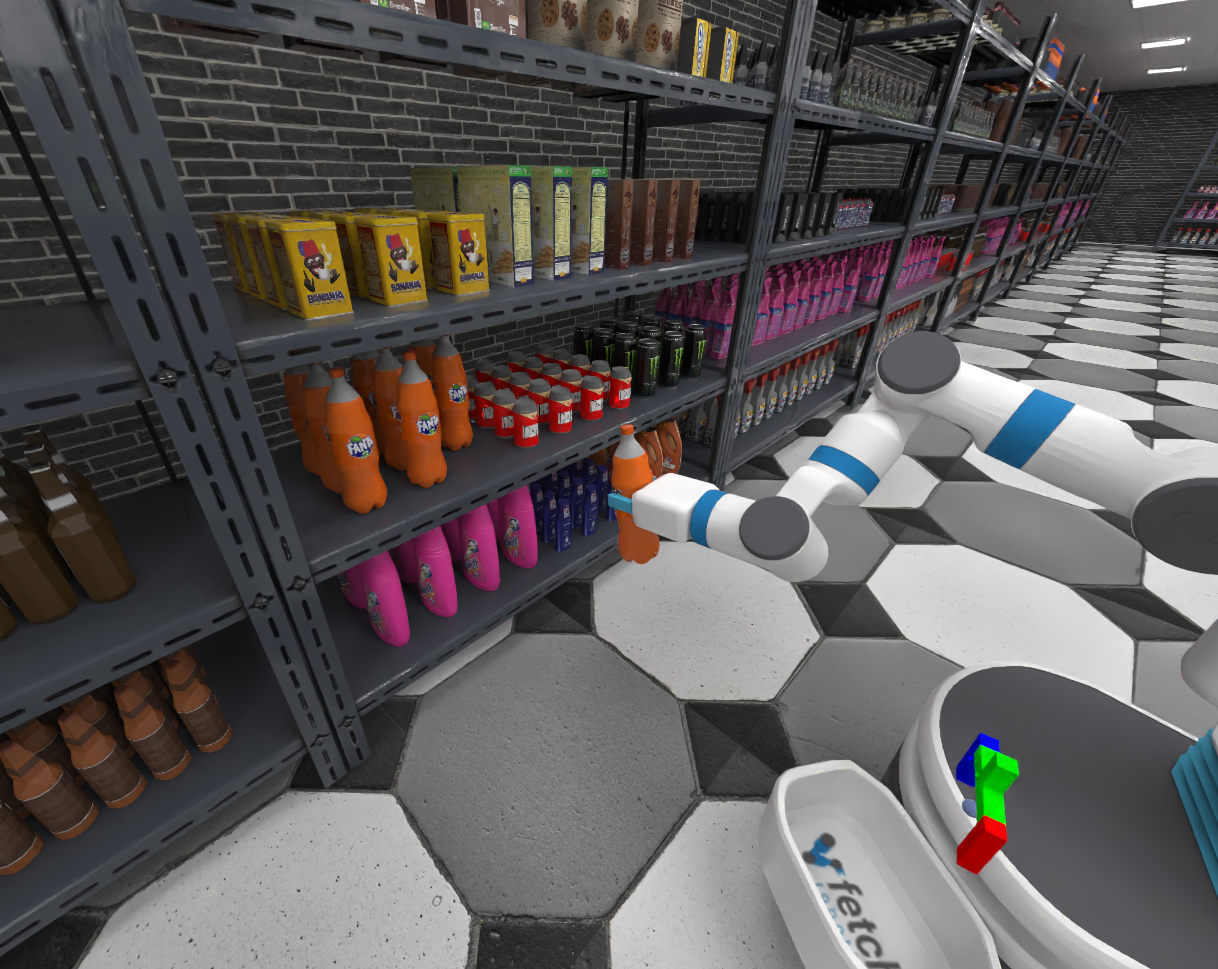}
    \end{subfigure}
    \caption{Examples of heuristically generated anchor poses used in our motion planner.}
    
    \label{fig:mp:anchors}
\end{figure*}

Store Trajectory Sampler leverage motion planning to generate trajectory data.
In our motion-planning pipeline, we collect successful mobile-manipulation demonstrations for each task.
For every task, we heuristically specify a sequence of initial, intermediate, and final anchor poses corresponding to the main stages of execution.
During data collection, these poses are randomized to increase the diversity of demonstrations.
An example sequence of anchor poses is shown in Figure~\ref{fig:mp:anchors}.

Motion planning is performed sequentially between consecutive anchor poses.
For segments that do not require mobile-base movement, we first attempt to generate a trajectory using screw motion~\cite{murray2017mathematical}.
Since screw motion does not explicitly account for obstacles during trajectory generation, the resulting trajectory is checked for collisions.
If it is invalid, we instead use RRT-Connect~\cite{kuffner2000rrt}, which explicitly reasons about scene obstacles.
For segments requiring mobile-base repositioning, we use task-specific heuristic planners designed to approach the relevant fixture and support subsequent manipulation.
If planning or execution fails at any stage, the attempt is discarded and the environment is reset for a new collection attempt.
Only trajectories that successfully complete the full task are retained in the demonstration dataset.

\section{Store Robotics Benchmark}

The goal of our benchmark is to evaluate the capabilities of current state-of-the-art generalist policies in retail environments, with fine-tuning using generated trajectories.
It is built on ManiSkill3~\cite{taomaniskill3}, a high-performance, high-fidelity robot simulation framework with realistic physics and ray-traced rendering.
We use the Fetch robot (Figure~\ref{fig:bench:pick-to-basket}), a mobile manipulator with a differential-drive base, a 7-DOF arm, and a prismatic torso joint for vertical reach. 
Its parallel gripper, combined with accurate contact modeling in ManiSkill3, enables versatile object manipulation in dense, realistic retail settings.

\subsection{Testing Scenarios}

To assess the generalization capabilities of 
policies fine-tuned within our benchmark, we consider the following axes of environment and task variation:
\begin{enumerate}
\item \textit{Robot Position}\label{en:rand:RP}: Randomized start positions within task-relevant regions.
Used during training trajectory generation.
\item \textit{Textures}\label{en:rand:T}: Random variations in wall, floor, ceiling, and door textures. 
Present during training.
\item \textit{Store Layout}\label{en:rand:SL}: Unseen store layouts at test time represent an out-of-distribution (OOD) domain.
\item \textit{Unseen Shelf Arrangement}\label{en:rand:USA}: Shelf arrangements not encountered during training.
\item \textit{Unseen in Task Items}\label{en:rand:UTI}: Items encountered during training in other tasks but not in the target task.
Considered OOD with respect to the specific task.
\item \textit{Completely Unseen Items}\label{en:rand:CUI}: Items not encountered during training at all. 
Represents a more challenging form of OOD.
\end{enumerate}
To balance benchmarking feasibility with meaningful generalization evaluation, we define the following testing scenarios:
\begin{itemize}
\item \textbf{In-Domain}: Robot position randomization only (\ref{en:rand:RP}).
\item \textbf{Unseen Scenes}: Robot position, texture, and store layout randomization (\ref{en:rand:RP} + \ref{en:rand:T} + \ref{en:rand:SL}).
\item \textbf{Unseen Scenes \& Items}: Unseen scenes with additional OOD items drawn from other tasks (\ref{en:rand:RP} + \ref{en:rand:T} + \ref{en:rand:SL} + \ref{en:rand:UTI}).
\end{itemize}
Although our benchmark supports scenarios \ref{en:rand:USA} and \ref{en:rand:CUI}, we exclude them from evaluation, as current policies consistently fail even under the simpler conditions considered above (see Table~\ref{tab:results}).

These configurations allow us to evaluate policy generalization across increasingly challenging deployment conditions, including novel scene layouts and object combinations.

\subsection{Tasks}

We design a set of atomic tasks covering fundamental manipulation skills required to accomplish retail-related objectives:
\begin{itemize}
    \item Pick to basket\,---\,the robot must pick an item from a shelf or refrigerator and place it into a cart.
    \item Pick from floor\,---\,an item that has fallen to the floor must be picked up and returned to its appropriate location.
    \item From board to board\,---\,the robot transfers an item from one board to another.
    \item Open fridge\,---\,the robot opens the door of a refrigerator.
    \item Close fridge\,---\,the robot closes the door of a refrigerator.
\end{itemize}
Building on atomic tasks, we define composite tasks:
\begin{itemize}
    \item Pick \{N\} items\,---\,the robot is provided with a list of N items and must collect all specified products and place them in the cart.
    \item Pick from fridge\,---\,open the fridge, pick an item, and close the fridge.
\end{itemize}
Each task includes a textual instruction specifying the target item and fixture names.
The robot is expected to interact with the nearest matching instances, as no specific objects are explicitly designated.
When assessing task completion, we evaluate not only the final positions of target products and fixtures, but also verify that surrounding items were not disturbed or collided with during policy execution.

\subsection{Generalist Baselines}

We evaluate four state-of-the-art generalist VLA models: lightweight transformer Octo~\cite{octo_2023}
and LLM-based SmolVLA~\cite{shukor2025smolvlavisionlanguageactionmodelaffordable}, $\pi_0$~\cite{black2024pi0visionlanguageactionflowmodel} and $\pi_{0.5}$~\cite{intelligence2025pi05visionlanguageactionmodelopenworld}.
Each model is fine-tuned via imitation learning using trajectories generated by our Store Trajectory Sampler.

To reduce computational overhead and model a standard, realistic VLA adaptation setup for a new domain and tasks, we generate only 248 trajectories per (task, item, fixture) triplet, for a total of 2,976 demonstrations.
To evaluate generalization, we restrict the number of training objects per task to 2–3 and ensure that these objects remain unseen in all other tasks.
Although our simulator supports a variety of item arrangements, we use fully packed shelves for both training and testing (Figure~\ref{fig:product-arrangement:day1}), as models already struggle under this simplified configuration.
Further details on the collected trajectories can be found 
in Appendix~\ref{app:dataset}.

Models are finetuned exclusively on atomic tasks.
Composite tasks are evaluated by decomposing them into sequences of atomic instructions, executed step by step using the same policy.
Further details on fine-tuning and hyperparameters are provided 
in Appendix~\ref{app:finetune}.

\subsection{Evaluation Results}

Evaluation results are presented in Table~\ref{tab:results}.
We evaluate performance on atomic and composite tasks using the mean success rate per task.
We use 100 trials per (task, item, fixture) triplet to estimate success rates.

\begin{table*}[!ht]
\centering
\resizebox{\columnwidth}{!}{%
\fontsize{9pt}{10pt}\selectfont
\tabcolsep=2.5pt
\begin{tabular}{cccccccccc}
\toprule
\multirow{3}{*}{Model} &
\multirow{3}{*}{Param. (\#)} &
\multirow{3}{*}{\begin{tabular}[c]{@{}c@{}}Testing\\ scenario\end{tabular}} &
\multicolumn{5}{c}{Atomic Tasks} &
\multicolumn{2}{c}{Composite Tasks} \\
\cmidrule(lr){4-8}
\cmidrule(lr){9-10}
& & &
\begin{tabular}[c]{@{}c@{}}Pick to\\ basket\end{tabular} &
\begin{tabular}[c]{@{}c@{}}Pick from \\ floor\end{tabular} &
\begin{tabular}[c]{@{}c@{}}From board\\ to board\end{tabular} &
\begin{tabular}[c]{@{}c@{}}Open \\ fridge\end{tabular} &
\begin{tabular}[c]{@{}c@{}}Close \\ fridge\end{tabular} &
\begin{tabular}[c]{@{}c@{}}Pick 3 \\ items\end{tabular} &
\begin{tabular}[c]{@{}c@{}}Pick from \\ fridge\end{tabular} \\
\midrule
\multirow{3}{*}{Octo} & \multirow{3}{*}{93M} & In-Domain   & 21 & 1 & 21 & 27 & 45 & 0  & 0\\
& & Unseen Scenes   & 2 & 1 & 1 & 23 & 26 & 0 & 0\\
& & Unseen Scenes \& Items  & 0 & 0 & 0 & \nmark & \nmark & 0  & 0\\
\midrule
\multirow{3}{*}{SmolVLA} & \multirow{3}{*}{450M} & In-Domain   & 0 & 0 & 0 & 10 & 13 & 0  & 0\\
& & Unseen Scenes   & 0 & 0 & 0 & 13 & 13 & 0 & 0\\
& & Unseen Scenes \& Items  & 0 & 0 & 0 & \nmark & \nmark & 0  & 0\\
\midrule
\multirow{3}{*}{$\pi_0$} & \multirow{3}{*}{3.3B} & In-Domain   & 14 & 21 & 19 & 61 & 95 & 0  & 0\\
& & Unseen Scenes   & 1  & 6  & 3  & 44  & 85 & 0  & 0\\
& & Unseen Scenes \& Items  & 0  & 1  & 0  & \nmark  & \nmark & 0  & 0\\
\midrule
\multirow{3}{*}{$\pi_{0.5}$} & \multirow{3}{*}{3.3B} & In-Domain   & 55 & 21 & 56 & 56 & 91 & 0  & 0\\
& & Unseen Scenes   & 31  & 4  & 27  & 49  & 78 & 0  & 0\\
& & Unseen Scenes \& Items  & 5  & 3  & 21  & \nmark  & \nmark & 0  & 0\\
\bottomrule
\end{tabular}}
\caption{Average success rates (\%) of generalist VLA models on atomic and composite retail tasks across different testing scenarios.
Higher values indicate better performance. 
\nmark{} indicate that scenario is not applicable for the task.
}
\label{tab:results}
\end{table*}

Moreover, to better understand the low task success rates, we annotate a subset of failed episodes according to the earliest primary failure in the manipulation pipeline (Appendix~\ref{app:eval}).
Our taxonomy includes high-level grounding errors (\textit{task}, \textit{target}), navigation and approach failures (\textit{move}, \textit{pregrasp}), grasping failures (\textit{grasp}), post-grasp side effects (\textit{drop}, \textit{displace}), placement failures (\textit{preplace}, \textit{place-coord}, \textit{place}), and incomplete articulation of a fridge door (\textit{partial}).
We report the resulting failure distribution for each model in Table~\ref{tab:failures_share}. 
Detailed definitions of the failure categories and additional failure analysis are provided in Appendix~\ref{app:eval}.

\begin{table*}[!ht]
\centering
\fontsize{9pt}{10pt}\selectfont
\tabcolsep=2.5pt
\begin{tabular}{lccccccccccc}
\toprule
 &
  \multicolumn{1}{l}{task} &
  \multicolumn{1}{l}{target} &
  \multicolumn{1}{l}{move} &
  \multicolumn{1}{l}{pregrasp} &
  \multicolumn{1}{l}{grasp} &
  \multicolumn{1}{l}{drop} &
  \multicolumn{1}{l}{displace} &
  \multicolumn{1}{l}{preplace} &
  \multicolumn{1}{l}{place-coord} &
  \multicolumn{1}{l}{place} &
  \multicolumn{1}{l}{partial} \\
\midrule
Octo        & 7 & {\ul 22} & 8           & \textit{9}  & \textbf{39} & 2 & 2 & 1 & 1 & 6          & 3 \\
SmolVLA     & 4 & {\ul 29} & \textit{16} & \textbf{32} & 17          & 0 & 0 & 0 & 0 & 0          & 2 \\
$\pi_0$     & 0 & {\ul 23} & 1           & \textit{9}  & \textbf{43} & 5 & 1 & 2 & 4 & 5          & 7 \\
$\pi_{0.5}$ & 0 & {\ul 20} & 1           & 2           & \textbf{50} & 4 & 7 & 0 & 2 & \textit{9} & 5 \\
\bottomrule
\end{tabular}
\caption{Rates of failures (\%) for each VLA model.
For each model, the highest failure rate is shown in \textbf{bold}, the second highest is {\ul underlined}, and the third highest is shown in \textit{italic}.
}
\label{tab:failures_share}
\end{table*}

The results in Table~\ref{tab:results} show that \textbf{current generalist VLA models struggle even with basic retail tasks}.
SmolVLA performs poorly across all scenarios, achieving non-zero success only on the simplest \textit{open fridge} and \textit{close fridge} tasks.
We hypothesize that this may be related to its more limited pretraining data: SmolVLA is pretrained on community-shared LeRobot datasets, whereas Octo, $\pi_0$, and $\pi_{0.5}$ are pretrained on larger and more diverse datasets.
Octo and $\pi_0$ achieve moderate performance in the In-Domain setting, but degrade in Unseen Scenes and fail completely in the Unseen Scenes \& Items setting.
$\pi_{0.5}$ performs significantly better and is the only model with non-zero success in Unseen Scenes \& Items, although its performance remains far from reliable.
All models achieve zero success on composite tasks, indicating limited ability to execute multi-step instructions and generalize across task stages.
Overall, these results indicate that the models’ relative rankings are consistent with their expected capabilities and with their performance on existing tabletop and household benchmarks.

The failure analysis in Table~3 reveals two main bottlenecks.
Octo, $\pi_0$, and $\pi_{0.5}$ fail predominantly at the \textit{grasp} stage, while SmolVLA most often fails during earlier \textit{pregrasp} positioning.
A second common issue is object grounding (\textit{target}), indicating difficulties in selecting the instructed product.
SmolVLA and Octo also exhibit failures in positioning the mobile base near the target fixture (\textit{move}).
Overall, current VLAs struggle primarily with precise shelf-level interaction and reliable product grounding in retail scenes.

Overall, these results highlight three limitations of current generalist models: \textbf{fragility to minor scene changes} (e.g., layouts, textures, object placements), \textbf{poor generalization from limited demonstrations} to novel object-task combinations, and \textbf{insufficient reliability for shelf-level grasping and compositional execution}.
Our findings suggest that existing \textbf{pretrained models may be insufficient} for effective adaptation in the retail domain, and that \textbf{targeted pretraining on retail-specific data may be necessary}.



\section{Limitations}

Despite providing a diverse retail simulation environment, \RBM{} has several limitations.
First, it currently supports only a parallel-jaw gripper, excluding suction-based end-effectors and dexterous hands.
Second, some wide or irregular packages cannot be grasped by the Fetch gripper. Therefore, we use relatively large inter-item gaps to enable reliable manipulation.
Third, the benchmark includes only rigid-body packages and omits deformable items.


\section{Conclusion}
\label{sec:conclusion}

In this work, we introduced a novel open-source \RBM{} suite for benchmarking robotic systems in retail environments, a domain with significant practical relevance and underexplored challenges.
The suite includes a procedural store layout generator, a trajectory generation pipeline, evaluation tools, and fine-tuned baseline models. 
Our experiments show that current state-of-the-art generalist models struggle with common retail scenarios, highlighting a clear mismatch between existing capabilities and the demands of real-world automation.
We hope our work will facilitate the development of more robust, scalable, and task-aware robotic systems.
All components of \RBM{} are publicly available to support future research in this direction.

\FloatBarrier

\bibliography{biblography}

\appendix

\section{Related Work}
\label{sec:related_work}

\subsection{Benchmarks and Datasets for Robotics}

Inspired by the success of large-scale pretraining in CV~\cite{He2015,dosovitskiy2020vit,pmlr-v139-radford21a,ravi2024sam2}
and NLP~\cite{devlin2019bertpretrainingdeepbidirectional,NEURIPS2020_1457c0d6,touvron2023llamaopenefficientfoundation},
robotics has pursued similar dataset development.
However, collecting diverse and scalable robotic data remains challenging due to platform heterogeneity and physical interaction requirements \cite{open_x_embodiment_rt_x_2023,khazatsky2024droid}.

Real-world evaluation is also difficult to standardize, often requiring human resets and suffering from environment variability. 
As a result, simulation-based benchmarks have become popular for their reproducibility and ease of use.

Existing benchmarks mostly focus on household tasks.
ALFRED~\cite{ALFRED20}, RLBench~\cite{james2019rlbench}, RoboCasa~\cite{robocasa2024} and ManiSkill-HUB~\cite{shukla2024maniskillhab} offer tasks involving navigation, manipulation, and instruction following.
Language-conditioned and lifelong learning are addressed in CALVIN~\cite{mees2022calvin}, LIBERO~\cite{liu2023libero}, VLABench~\cite{zhang2024vlabench}, and BEHAVIOR-1K~\cite{li2022behavior}.

However, retail and logistics scenarios\,---\,such as shelf picking or order packing\,---\,remain underexplored. 
Dedicated benchmarks for these domains are needed to advance robotic capabilities in retail environments.

\subsection{Retail Domain}

A number of datasets have been developed targeting retail-related computer vision tasks, including product classification~\cite{peng2020rp2k}, product detection~\cite{goldman2019precise,lindermayr2023ipa}, change detection and depth estimation~\cite{mata2022standardsim}.

Real-world datasets such as SKU110K~\cite{goldman2019precise} and RP2K~\cite{peng2020rp2k} are valuable for pretraining and evaluating the perception modules of robotic systems.
However, they contain only 2D images of products and shelves, making them unsuitable for training or benchmarking robotic manipulation and navigation.

IPA-3D1K~\cite{lindermayr2023ipa} includes 1,000 high-quality 3D assets of real retail products, but it is not yet publicly available. 
The synthetic dataset StandardSim~\cite{mata2022standardsim} offers only 2D image data, generated using 456 purchased product assets and a limited set of store layouts.
Neither dataset provides code for generating scenes, product arrangements, or robotic trajectories, limiting their applicability to end-to-end robotics research.

FetchBot~\cite{liu2025fetchbot} introduces the UniVoxGen method for fast generation of highly cluttered shelf arrangements.
However, it focuses solely on atomic picking task and does not generate full store layouts or visualize product textures, which limits its applicability for benchmarking end-to-end robotic policies for retail.

In our work, we address these limitations by providing code to generate diverse store layouts and robotic trajectories, enabling the training and benchmarking of robotic policies in retail environments.

\subsection{Trajectories Generation}


Collecting robot trajectories via human teleoperation is costly and time-consuming~\cite{liu2023libero,zhang2024vlabench}, while simulators offer privileged access to object states. 
As a result, automatic trajectory generation has become increasingly important.

Motion planning methods (e.g., RRT*~\cite{KaramanSamplingBased}, CHOMP~\cite{ZuckerChomp}) can generate collision-free paths but require manual task specification, limiting their scalability.
Reinforcement learning (RL)~\cite{sutton2018reinforcement} learns from rewards~\cite{haarnoja2017soft,andrychowicz:hal-03162554}, but designing suitable reward functions is challenging, training is computationally intensive, and resulting behaviors may be suboptimal.

A promising alternative is demonstration-based augmentation.
Methods like MimicGen~\cite{mandlekar2023mimicgen} scale a small set of human demonstrations by adapting them to new scenes, enabling diverse and reusable trajectory generation.

In our work, we adapt motion planning methods to generate trajectories in retail store environments.

\subsection{Robotic Models}

Recent generalist models aim to unify perception and control across diverse tasks and robots.
Octo~\cite{octo_2023}, trained on 800k trajectories from 9 platforms, uses a transformer-based policy conditioned on goals via language or images.
It generalizes well and can be quickly adapted to new embodiments.
OpenVLA~\cite{kim24openvla}, a 7B Vision-Language-Action model, outperforms larger models like RT-2 on 29 manipulation tasks using multi-view visual features and a Llama-2 backbone. 
It supports efficient fine-tuning and strong multi-object reasoning.
Pi0~\cite{black2024pi0visionlanguageactionflowmodel} combines a vision-language encoder with a flow-matching policy head.
Trained on 68 tasks across multiple robots, it adapts to new tasks with minimal data and supports varied embodiments, including mobile and dual-arm robots.

Despite strong results on household and tabletop tasks, these models remain untested in retail logistics.
Real-world scenarios like warehouse picking or packing involve larger spaces, dynamic layouts, cluttered environments and time-critical demands.
A dedicated benchmark is needed to evaluate how well such generalist policies transfer to retail environments.

\section{Societal Impact and Ethical Considerations}

We develop \RBM{} as a simulated benchmark to study and evaluate robotic automation in retail (dark-store) environments.
Automating repetitive picking and restocking can reduce physical strain and injury risk for workers, enable more reliable and scalable 24/7 order fulfillment, and improve the efficiency and accessibility of delivery services, including for customers with limited mobility or in underserved areas.

At the same time, increased automation raises concerns about job displacement, changes in working conditions, and greater reliance on monitoring and data collection in retail spaces.
While \RBM{} itself is purely a simulated environment and does not directly involve workers or customers, we acknowledge that advances enabled by research using \RBM{} may influence how retail work is organized in the future.

\section{Tasks and Datasets}
\label{app:dataset}

\begin{table*}[!bht]
    \centering
    \resizebox{\columnwidth}{!}{%
    \fontsize{9pt}{10pt}\selectfont
    \tabcolsep=2pt
    \begin{tabular}{lllll}
    \toprule
        Task & Skill Family & Description & Success Criteria & Language instruction example \\ \midrule
        Pick to basket & Pick and place &
        \begin{tabular}[c]{@{}l@{}l@{}}Pick an object with specified \\name from the shelf and\\ place it in the basket.\end{tabular}
        & \begin{tabular}[c]{@{}l@{}l@{}}Any object of the target type\\ is inside the basket, other items\\ are not moved, the robot is static.\end{tabular}
        & \begin{tabular}[c]{@{}l@{}l@{}}\texttt{move to the shelf,} \\ \texttt{pick the fanta bottle,} \\\texttt{and place it in} \\ \texttt{the basket}\end{tabular}\\ 
        \addlinespace[1em]

        Pick from floor & Pick and place &
        \begin{tabular}[c]{@{}l@{}l@{}}Pick an object from the \\ floor and place it to the \\ target shelf (the second \\ or the third board). \end{tabular}
        & \begin{tabular}[c]{@{}l@{}l@{}}The object is placed near the \\ correct group of products, other \\ items are not moved, and \\ the robot remains static.\end{tabular}
        & \begin{tabular}[c]{@{}l@{}l@{}}\texttt{pick the SLAM luncheon} \\ \texttt{meat from the floor and} \\\texttt{place it on the shelf}\end{tabular}\\
        \addlinespace[1em]
        
        \begin{tabular}[c]{@{}l@{}l@{}}From board to \\ board \end{tabular} & Pick and place &
        \begin{tabular}[c]{@{}l@{}l@{}}Pick an object with the \\ specified name from one \\ board and place it one board \\ higher (from the second \\ board to the third, or from \\ the third to the fourth). The \\ target board is empty. \end{tabular}
        & \begin{tabular}[c]{@{}l@{}l@{}}The object is placed near the \\ correct group of products, other \\ items are not moved, \\ the robot is static.\end{tabular}
        & \begin{tabular}[c]{@{}l@{}l@{}}\texttt{pick the Duff Beer Can} \\ \texttt{and place it} \\ \texttt{on an empty board} \end{tabular}\\
        \midrule
        Open showcase & \begin{tabular}[c]{@{}l@{}l@{}} Open/close \end{tabular} &
        \begin{tabular}[c]{@{}l@{}l@{}} Open one of the four doors \\ of the vertical showcase. \\ The doors are named from \\ left to right as ``first'', \\ ``second'', ``third'', ``fourth''. \end{tabular}
        & \begin{tabular}[c]{@{}l@{}l@{}}The specified door is opened, \\ the robot is static.\end{tabular}
        & \begin{tabular}[c]{@{}l@{}l@{}}\texttt{open the second} \\ \texttt{door of the showcase}\end{tabular}\\

        \addlinespace[1em]

        Close showcase & \begin{tabular}[c]{@{}l@{}l@{}} Open/close \end{tabular} &
        \begin{tabular}[c]{@{}l@{}l@{}} Close one of the four doors \\ of the vertical showcase \\ that is already open. \end{tabular}
        & \begin{tabular}[c]{@{}l@{}l@{}}The specified door is closed, \\ the robot is static.\end{tabular}
        & \begin{tabular}[c]{@{}l@{}l@{}}\texttt{close the door} \\ \texttt{of the showcase}\end{tabular}\\

        \addlinespace[1em]

        Open fridge & \begin{tabular}[c]{@{}l@{}l@{}} Open/close \end{tabular} &
       \begin{tabular}[c]{@{}l@{}l@{}} Open the door of the small \\ horizontal ice cream fridge. \end{tabular}
        & \begin{tabular}[c]{@{}l@{}l@{}}The door is open,\\the robot is static.\end{tabular}
        & \begin{tabular}[c]{@{}l@{}l@{}}\texttt{open the fridge}\end{tabular}\\

        \addlinespace[1em]
        
         Close fridge & \begin{tabular}[c]{@{}l@{}l@{}} Open/close \end{tabular} &
       \begin{tabular}[c]{@{}l@{}l@{}} Close the door of the small \\ horizontal ice-cream fridge \\ that is already open. \end{tabular}
        & \begin{tabular}[c]{@{}l@{}l@{}}The door is closed,\\ the robot is static.\end{tabular}
        & \begin{tabular}[c]{@{}l@{}l@{}}\texttt{close the fridge}\end{tabular}\\
         
        \bottomrule
    \end{tabular}}
    \caption{Atomic tasks descriptions}
    \label{tab:tasks}
\end{table*}

\begin{table*}[!bht]
    \centering
    \fontsize{9pt}{10pt}\selectfont
    \tabcolsep=2pt
    \begin{tabular}{llll}
    \toprule
         & Pick to basket & \begin{tabular}[c]{@{}l@{}l@{}}From board to \\ board \end{tabular} & Pick from floor \\ 
         \midrule
         
        Train items & \begin{tabular}[c]{@{}l@{}l@{}} 
         Nivea Body Milk \\ Nestle Honey Stars \\ Fanta \end{tabular} & \begin{tabular}[c]{@{}l@{}l@{}} Nestle Cereals \\ Duff Beer Can \\ Vanish \end{tabular}
        & \begin{tabular}[c]{@{}l@{}l@{}} Heinz Beans \\ SLAM luncheon meat \end{tabular} \\

        \midrule
        
        Test items & \begin{tabular}[c]{@{}l@{}l@{}} Nestle Cereals \\ SLAM luncheon meat \end{tabular} & \begin{tabular}[c]{@{}l@{}l@{}} Nivea Body Milk \\ Fanta \end{tabular}
        & \begin{tabular}[c]{@{}l@{}l@{}} Vanish \\ Duff Beer Can \end{tabular} \\
         
        \bottomrule
    \end{tabular}
    \caption{Train/test items split.}
    \label{tab:item_distr}
\end{table*}

\subsection{Atomic Tasks}

We define five atomic tasks encompassing two core manipulation skills: (1) \textit{pick and place}, (2) \textit{open/close} doors. 
Task details are outlined in Table~\ref{tab:tasks}.
At the start of each episode, the robot's initial pose is randomized near the target shelf or fridge.

To assess generalization in \textit{pick and place} tasks, we selected a subset of eight product items and partitioned them into train/test splits.
For each task, test items were excluded from training on that specific task but were present in the training set of other tasks. See Table~\ref{tab:item_distr} for the full item distribution.

\subsection{Composite Tasks}

To evaluate the models' ability to execute multiple atomic tasks sequentially, we designed two long-horizon tasks.
The first, \textit{Pick 3 items}, requires the robot to perform the \textit{Pick to basket} task for three different items in sequence. 
The second task, \textit{Pick from fridge}, simulates a scenario where the robot must retrieve an item from a closed refrigerated showcase.
This task comprises three atomic subtasks: \textit{Open fridge}, \textit{Pick to basket}, and \textit{Close fridge}.

We evaluate policies in the long-horizon setup using an oracle that decomposes each composite task into a sequence of atomic subtasks, executed sequentially upon the successful completion of the preceding subtask.

\subsection{Training Dataset}

To obtain demonstration trajectories we employed motion planning algorithms from \texttt{mplib}\footnote{https://motion-planning-lib.readthedocs.io/latest/} package.
We collected 248 demonstration trajectories per train object for \textit{pick and place} tasks and 248 per \textit{open/close} tasks.
A total of 2,976 trajectories were collected, comprising 1,401,169 transitions.
The whole process of data generation (scene synthesis, motion planning and camera rendering) takes approximately 8 hours on an NVIDIA V100 GPU.

Our training data comprises the following components:
\begin{enumerate}
    \item \textbf{Observations}, including:
        \begin{itemize}
        \item \textbf{Textual command}: describes the task to perform along with the target objects.
        \item \textbf{Images}: RGB views from the left shoulder camera (256×256×3), gripper camera (128×128×3), and right shoulder camera (256×256×3).
Note that we did not employ images captured from the native Fetch head camera, as this view is usually blocked by the robot’s hand.
        \item \textbf{Proprioception}: joint positions and velocities.
\end{itemize}
    \item \textbf{Actions}, represented as 11-dimensional vectors:
    \begin{itemize}
        \item 7 values for arm joint positions,
        \item 1 for gripper control,
        \item 1 for vertical torso motion,
        \item 2 for base control: forward/backward linear velocity and rotational velocity around the vertical axis.
    \end{itemize}
\end{enumerate}

We use the Proportional-Derivative (PD) joint position target control mode (referred to as \texttt{pd\_joint\_pos} in ManiSkill3) to control the robot joints, with the exception of the robot base, which is controlled by specifying the linear and angular velocities. 
The simulator uses these target positions in combination with a PD controller to compute the torques required to move the joints to the desired positions.
Also, we changed joints limits since our motion planning library does not support continuous joints.

\section{Baselines Fine-tuning}
\label{app:finetune}

\subsection{Octo}

We employed the official JAX implementation of Octo\footnote{https://github.com/octo-models/octo}.
We fully finetuned the model following the provided scripts.
The model was trained for 1M iterations with a batch size of 128 on 8 A100 GPUs. 
We finetuned it using 2 history observations and a continuous action head with the action horizon of length 50. All other settings remain at their default values.
The finetuning process took approximately 3 days.

\subsection{SmolVLA}

We used the official Pytorch implementation of SmolVLA\footnote{https://github.com/huggingface/lerobot}.
We fully finetuned the model following the provided scripts. The model was trained to convergence for 200,000 iterations with a batch size of 64 on a single A100 GPU.
We finetuned it using an action horizon of length 50. All remaining settings use default values. The finetuning process took around 4 days.

\subsection{$\pi_0$ and $\pi_{0.5}$}

We use the official JAX implementations\footnote{https://github.com/Physical-Intelligence/openpi} and finetune $\pi_0$ and $\pi_{0.5}$ on our dataset using the provided finetuning scripts. 
We opt for full finetuning starting from the provided checkpoints.
The AdamW optimizer is used, with the learning rate following a cosine decay schedule, starting at a peak learning rate of 2.5e-5 and decaying to 2.5e-6. The models are trained to convergence for 500,000 steps on 8xA100 GPUs. The batch size is set to 256, and the action horizon is 50. All remaining settings use default values. The finetuning process took approximately 8 days.

\begin{table*}[!tbh]
\centering
\resizebox{\columnwidth}{!}{%
\fontsize{9pt}{10pt}\selectfont
\tabcolsep=2.5pt
\begin{tabular}{ccccccccccccc}
\toprule
\multirow{2}{*}{Eval mode} & \multicolumn{3}{c}{From board to board} & \multicolumn{2}{c}{Open} & \multicolumn{2}{c}{Close} & \multicolumn{2}{c}{Pick from floor} & \multicolumn{3}{c}{Pick to basket} \\
\cmidrule(lr){2-4} \cmidrule(lr){5-6} \cmidrule(lr){7-8} \cmidrule(lr){9-10} \cmidrule(lr){11-13}
& duff & nestle & vanish & fridge & showc. & fridge & showc. & heinz & slam & fanta & nivea & stars \\
\midrule
Train Seeds & 58 & 35 & 30 & 46 & 3 & 70 & 51 & 58 & 70 & 37 & 57 & 49 \\
In-Domain & 30 & 16 & 17 & 54 & 0 & 59 & 30 & 2 & 0 & 18 & 26 & 19 \\
Unseen Scenes & 2 & 1 & 0 & 45 & 0 & 43 & 9 & 1 & 1 & 1 & 2 & 2 \\
\bottomrule
\end{tabular}}
\caption{Average success rates (\%) of the generalist VLA model Octo on atomic retail tasks across different testing scenarios.
Higher values indicate better performance. 
}
\label{tab:octo_res}
\end{table*}

\begin{table*}[!tbh]
\centering
\resizebox{\columnwidth}{!}{%
\fontsize{9pt}{10pt}\selectfont
\tabcolsep=2.5pt
\begin{tabular}{ccccccccccccc}
\toprule
\multirow{2}{*}{Eval mode} & \multicolumn{3}{c}{From board to board} & \multicolumn{2}{c}{Open} & \multicolumn{2}{c}{Close} & \multicolumn{2}{c}{Pick from floor} & \multicolumn{3}{c}{Pick to basket} \\
\cmidrule(lr){2-4} \cmidrule(lr){5-6} \cmidrule(lr){7-8} \cmidrule(lr){9-10} \cmidrule(lr){11-13}
& duff & nestle & vanish & fridge & showc. & fridge & showc. & heinz & slam & fanta & nivea & stars \\
\midrule
Train Seeds & 0 & 0 & 0 & 30 & 0 & 31 & 4 & 0 & 0 & 0 & 0 & 0 \\
In-Domain & 0 & 0 & 0 & 19 & 0 & 25 & 1 & 0 & 0 & 0 & 0 & 0 \\
Unseen Scenes & 0 & 0 & 0 & 25 & 0 & 23 & 2 & 0 & 0 & 0 & 0 & 0 \\
\bottomrule
\end{tabular}}
\caption{Average success rates (\%) of the generalist VLA model SmolVLA on atomic retail tasks across different testing scenarios.
Higher values indicate better performance. 
}
\label{tab:smol_res}
\end{table*}

\begin{table*}[!tbh]
\centering
\resizebox{\columnwidth}{!}{%
\fontsize{9pt}{10pt}\selectfont
\tabcolsep=2.5pt
\begin{tabular}{ccccccccccccc}
\toprule
\multirow{2}{*}{Eval mode} & \multicolumn{3}{c}{From board to board} & \multicolumn{2}{c}{Open} & \multicolumn{2}{c}{Close} & \multicolumn{2}{c}{Pick from floor} & \multicolumn{3}{c}{Pick to basket} \\
\cmidrule(lr){2-4} \cmidrule(lr){5-6} \cmidrule(lr){7-8} \cmidrule(lr){9-10} \cmidrule(lr){11-13}
& duff & nestle & vanish & fridge & showc. & fridge & showc. & heinz & slam & fanta & nivea & stars \\
\midrule
Train Seeds & 47 & 22 & 20 & 96 & 38 & 100 & 96 & 55 & 52 & 34 & 39 & 28 \\
In-Domain & 33 & 11 & 14 & 95 & 26 & 100 & 90 & 26 & 16 & 13 & 17 & 13 \\
Unseen Scenes & 7 & 1 & 2 & 85 & 2 & 96 & 74 & 9 & 2 & 1 & 1 & 1 \\
\bottomrule
\end{tabular}}
\caption{Average success rates (\%) of the generalist VLA model $\pi_0$ on atomic retail tasks across different testing scenarios.
Higher values indicate better performance. 
}
\label{tab:pi0_res}
\end{table*}

\begin{table*}[!tbh]
\centering
\resizebox{\columnwidth}{!}{%
\fontsize{9pt}{10pt}\selectfont
\tabcolsep=2.5pt
\begin{tabular}{ccccccccccccc}
\toprule
\multirow{2}{*}{Eval mode} & \multicolumn{3}{c}{From board to board} & \multicolumn{2}{c}{Open} & \multicolumn{2}{c}{Close} & \multicolumn{2}{c}{Pick from floor} & \multicolumn{3}{c}{Pick to basket} \\
\cmidrule(lr){2-4} \cmidrule(lr){5-6} \cmidrule(lr){7-8} \cmidrule(lr){9-10} \cmidrule(lr){11-13}
& duff & nestle & vanish & fridge & showc. & fridge & showc. & heinz & slam & fanta & nivea & stars \\
\midrule
Train Seeds & 92 & 60 & 63 & 100 & 35 & 100 & 96 & 86 & 90 & 82 & 89 & 82 \\
In-Domain & 71 & 53 & 44 & 100 & 11 & 95 & 87 & 22 & 20 & 48 & 60 & 58 \\
Unseen Scenes & 45 & 23 & 12 & 98 & 0 & 97 & 59 & 8 & 0 & 37 & 36 & 19 \\
\bottomrule
\end{tabular}}
\caption{Average success rates (\%) of the generalist VLA model $\pi_{0.5}$ on atomic retail tasks across different testing scenarios.
Higher values indicate better performance. 
}
\label{tab:pi05_res}
\end{table*}

\begin{table*}[!tbh]
\centering
\tabcolsep=2.5pt
\begin{tabular}{cccccccc}
\toprule
\multirow{2}{*}{Eval mode} 
    & \multicolumn{2}{c}{From board to board} 
    & \multicolumn{2}{c}{Pick from floor} 
    & \multicolumn{2}{c}{Pick to basket} 
    & \multirow{2}{*}{ } \\
\cmidrule(lr){2-3}
\cmidrule(lr){4-5}
\cmidrule(lr){6-7}
& nivea & fanta & duff & fanta & nestle & slam \\
\midrule
Unseen Scenes \& Items  & 34 & 7 & 3 & 2 & 7 & 2 \\
\bottomrule
\end{tabular}
\caption{Average success rates (\%) of the generalist VLA model $\pi_{0.5}$ on atomic retail tasks for the Unseen Scenes \& Items scenario. Higher values indicate better performance.}
\label{tab:pi05_ood_res}
\end{table*}

\section{Evaluation and Failure Modes}
\label{app:eval}

\paragraph{Additional Evaluations.}
In Table~\ref{tab:octo_res}, we report detailed results of the Octo model evaluation for all tasks and objects. Table~\ref{tab:smol_res} presents the results for SmolVLA, Table~\ref{tab:pi0_res} shows detailed results of the $\pi_0$ model, and Table~\ref{tab:pi05_res} reports the $\pi_{0.5}$ results.
In addition to the ``In-Domain'' and ``Unseen Scenes'' scenarios, we also report success rates for the ``Train Seeds'' scenario, where the models are evaluated on the environments used during training.
Table~\ref{tab:pi05_ood_res} shows results of the $\pi_{0.5}$ model for the ``Unseen Scenes \& Items'' scenario with task-object combinations not seen during training. We do not report analogous tables for the other models as they fail completely in this scenario.
We also do not report results for composite tasks, as all models fail in this setup, as shown in the main paper.

\begin{table*}[!tbh]
\centering
\fontsize{9pt}{11pt}\selectfont
\tabcolsep=2.5pt
\begin{tabular}{ll}
\toprule
Tag & Definition \\
\midrule
\textit{task} & The robot attempted a different task than the one specified in the instruction. \\
\textit{target} & The robot acted on the wrong target object (one not named in the instruction). \\
\textit{move} & The robot failed to move its base to the required location. \\
\textit{pregrasp} & The robot failed to position itself correctly for grasping. \\
\textit{grasp} & \begin{tabular}[t]{@{}l@{}l@{}} The gripper is near the object but fails to close on it. For the fridge and showcase, \\ "object" refers to the door handle. \end{tabular}\\
\textit{drop} & The target object slipped out of the gripper after being grasped. \\
\textit{displace} & Non-target objects were displaced during execution. \\
\textit{preplace} & The robot failed to position itself correctly for placing the object. \\
\textit{place-coord} & The placement coordinates were chosen incorrectly. \\
\textit{place} & The robot is holding the object near the correct location but failed to release it into place. \\
\textit{partial} & The fridge sliding door or the showcase door was only partially opened or closed. \\
\bottomrule
\end{tabular}
\caption{Failure categories used for analysis of unsuccessful episodes.
Failures definitions, ordered by the stage of the manipulation pipeline at
which the failure occurs (high-level task selection $\xrightarrow{}$ navigation $\xrightarrow{}$ grasp $\xrightarrow{}$
post-grasp side effects $\xrightarrow{}$ placement).
}
\label{tab:failure-categories}
\end{table*}

\paragraph{Failure Analysis.} 
To analyze why evaluated policies fail in retail tasks, for each VLA model, we manually annotate a random subset of $\sim$10 unsuccessful episodes for each (task, item, fixture) triplet with the earliest primary failure observed in the manipulation pipeline (see Table~\ref{tab:failure-categories}).
In total we annotated 1581 failed episodes from ``Train Seeds'' (TS), ``In-Domain'' (ID),``Unseen Scenes'' (US) and ``Unseen Scenes \& Items'' (US\&I) scenarios.
The categories are ordered according to the stage at which a failure occurs, from high-level task interpretation to mobile-base positioning, grasping, post-grasp effects, and placement.

\begin{figure*}[tbh!]
    \centering
    \includegraphics[width=0.9\linewidth]{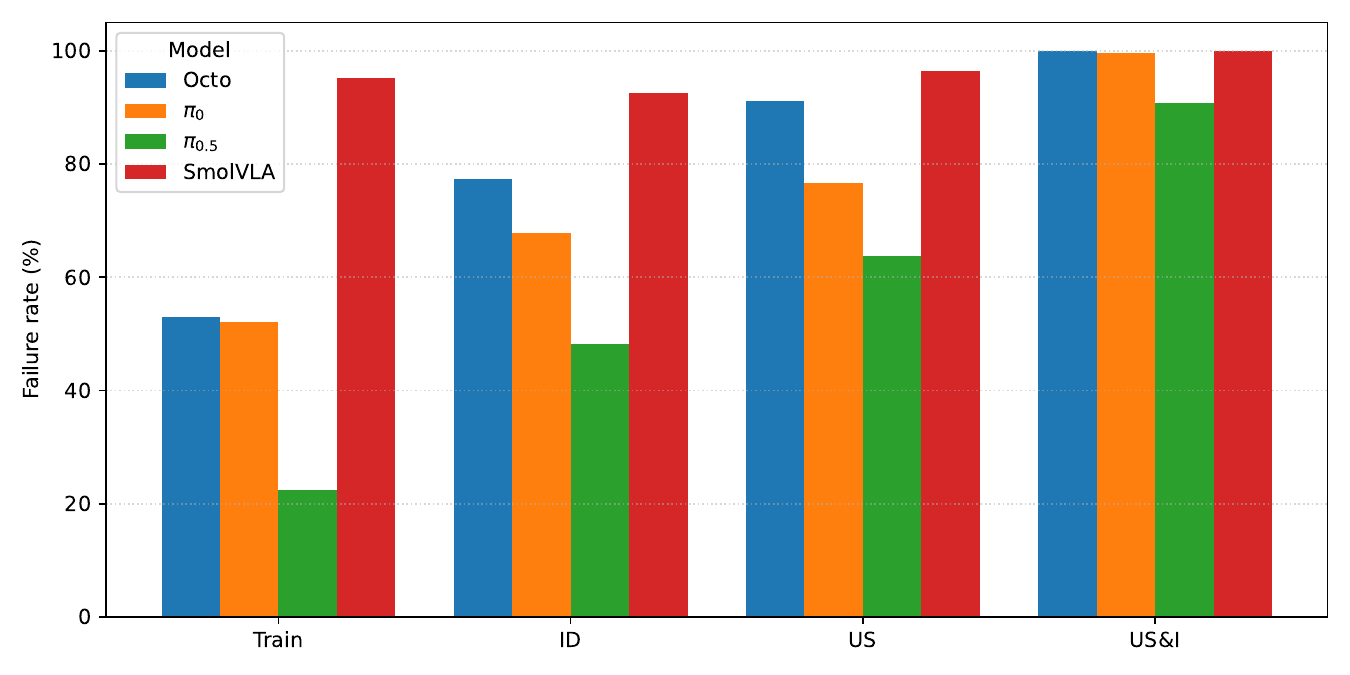}
    \caption{Failure rates (\%) of the evaluated VLA models across the \textit{Train Seeds} (TS), \textit{In-Domain} (ID), \textit{Unseen Scenes} (US), and \textit{Unseen Scenes \& Items} (US\&I) scenarios. Lower is better.}
    \label{fig:failure-rates}
\end{figure*}

We present additional analysis of VLA models failures.
Figure~\ref{fig:failure-rates} shows that failure rates increase substantially as the evaluation distribution moves farther from the training setting.
Among the evaluated models, $\pi_{0.5}$ is consistently the most robust, but its failure rate still becomes high in Unseen Scenes and exceeds 90\% in Unseen Scenes \& Items.
Octo and $\pi_0$ deteriorate sharply under scene and item shifts, reaching near-complete failure in the most challenging setting, while SmolVLA fails in almost all scenarios.
These results demonstrate that current generalist VLAs remain highly sensitive to retail-specific distribution shifts, even after adaptation.

\begin{figure*}[tbh!]
    \centering
    \includegraphics[width=\linewidth]{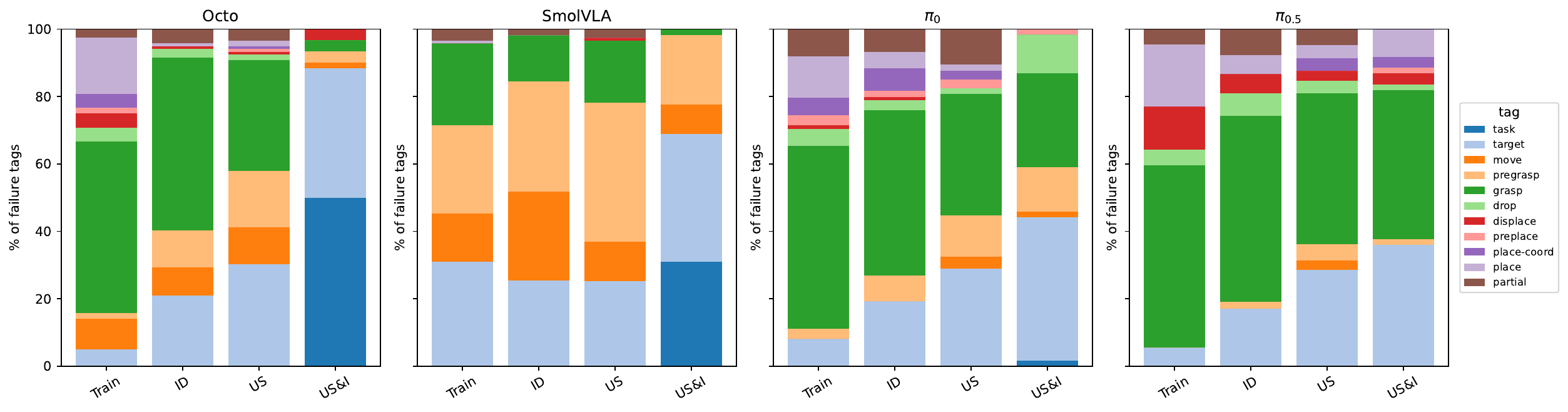}
    \caption{Distribution of primary failure types for each VLA model across the \textit{Train Seeds} (TS), \textit{In-Domain} (ID), \textit{Unseen Scenes} (US), and \textit{Unseen Scenes \& Items} (US\&I) scenarios. 
    Each failed episode is assigned the earliest primary failure in the manipulation pipeline; bars sum to 100\%.}
    \label{fig:failure-distributions}
\end{figure*}

Figure~\ref{fig:failure-distributions} shows that failure modes vary across models and evaluation scenarios.
In the less shifted settings, failures of Octo, $\pi_0$, and $\pi_{0.5}$ are largely dominated by unsuccessful grasping, while SmolVLA exhibits substantial pregrasp positioning failures.
Under stronger distribution shifts, especially in US\&I, Octo, $\pi_0$, and SmolVLA increasingly fail at earlier stages, such as task/target grounding and pregrasp positioning.
In contrast, $\pi_{0.5}$ more consistently reaches the interaction stage, but still frequently fails during grasping.
The results indicate that retail performance is limited by both grounding under distribution shift and reliable shelf-level manipulation.

\begin{figure*}[tbh!]
    \centering
    \includegraphics[width=\linewidth]{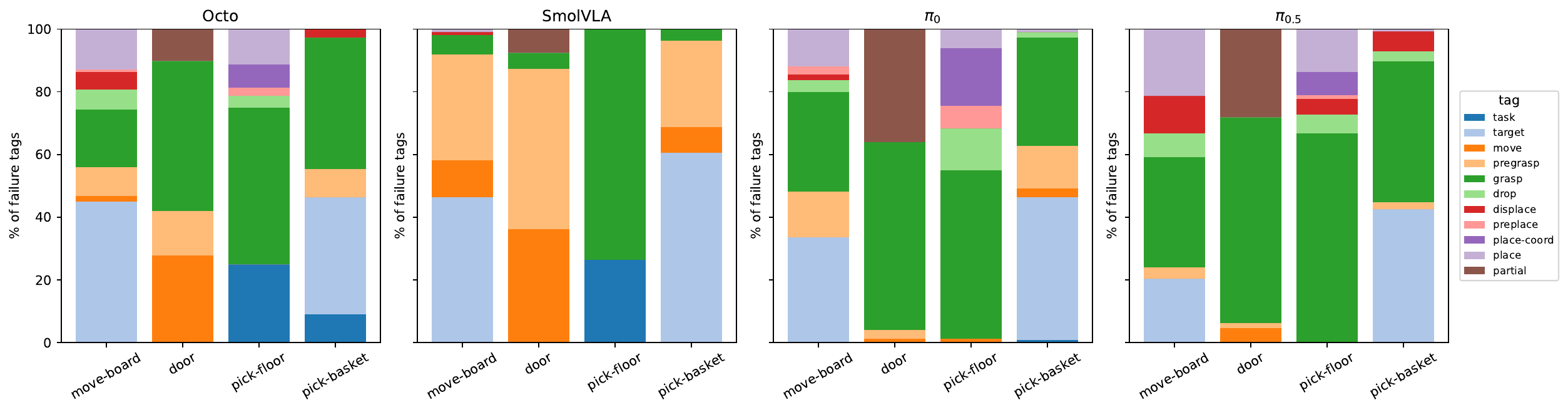}
    \caption{Distribution of primary failure categories for each VLA model across atomic tasks. 
    Here, \textit{move-board} denotes the \textit{From board to board} task;
    \textit{door} --- the door opening/closing tasks;
    \textit{pick-floor} --- \textit{Pick from floor};
    and \textit{pick-basket} --- \textit{Pick to basket}.
    Each failed episode is assigned its earliest primary failure in the manipulation pipeline; bars sum to 100\%.}
    \label{fig:failure-by-task}
\end{figure*}

Figure~\ref{fig:failure-by-task} shows that the main bottlenecks across atomic tasks are target grounding and grasping.
The door opening/closing tasks additionally expose articulation-specific failures, such as incomplete opening or closing.
Overall, retail tasks challenge VLAs both in selecting and manipulating target products and in interacting with articulated fixtures.


\section{Additional Simulation Optimizations}\label{sec:sup:sim_opt}

\begin{wrapfigure}{r}{0.5\textwidth}
  \centering
    \includegraphics[width=\linewidth]{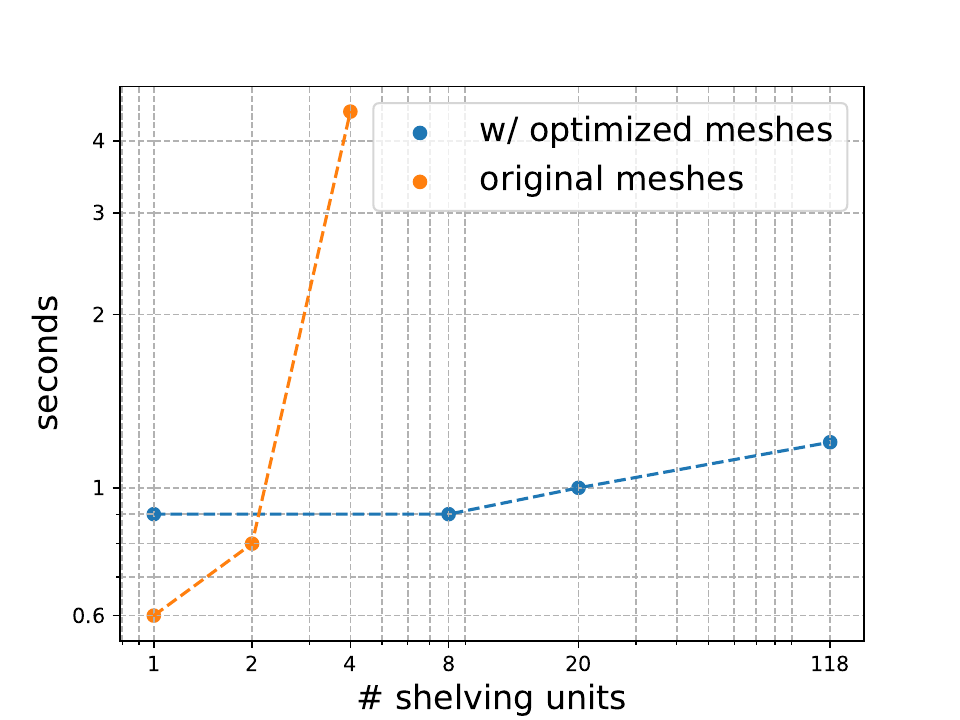}
    \caption{Simulation time for scenes with varying numbers of shelving units arranged with grocery items, comparing items optimized meshes (blue) to original meshes.}
    \label{fig:sup:perf}
\end{wrapfigure}
Inspired by hierarchical geometric models in computer graphics~\cite{clark1976hierarchical}, we optimize rendering performance in large-scale store simulations by representing items on shelving units, that are unlikely to be closely observed by the robot, using low-polygon assets obtained through the mesh optimization process described in Section~\ref{sec:assets}.

To evaluate the performance gains from our optimization, we conducted the following experiment.
We measured simulation time of 50 steps across store scenes with varying numbers of shelving units arranged with various products.
In the baseline setup, all shelves used original high-resolution product meshes.
In the optimized setup, only the shelf nearest to the robot used original meshes, while all background shelves used downscaled meshes.
As shown in Figure~\ref{fig:sup:perf}, optimized scenes yield substantial speedups\,---\,for example, simulating 120 shelves with optimized meshes is over three times faster than simulating just four shelves with unoptimized ones.
Experiments were conducted on Intel Xeon Gold 6278C CPU and NVIDIA V100 GPU.

Upon visual inspection of multiple rendered scenes from both the ego-view and human camera perspective, we observed that distant geometries appear visually indistinguishable regardless of asset detail level. 
This supports the use of low-polygon approximations to improve rendering speed and simulation efficiency.
Incorporating dynamic level-of-detail adjustment based on robot or camera pose and field of view is a promising direction for future improvements to the benchmark.

\section{Performance Benchmarking}

Below we report overall simulation performance of RoboBenchMart scenes with optimizations mentioned in Appendix~\ref{sec:sup:sim_opt}.
More specifically, we show FPS vs number of environments, annotated by GPU memory usage in GB on top of data points.
By FPS we mean the number of rendering and simulation steps made by several environments in parallel, following official ManiSkill documentation\footnote{https://maniskill.readthedocs.io/en/latest/user\_guide/ additional\_resources/performance\_benchmarking.html}:
\begin{equation*}
    \text{FPS}=\frac{\text{total\_steps} \cdot \text{num\_parallel\_envs}}{\text{elapsed\_time}}
\end{equation*}

\begin{figure*}[tbh!]
    \centering
    \begin{minipage}[t]{0.45\textwidth}
        \centering
        \includegraphics[width=\linewidth]{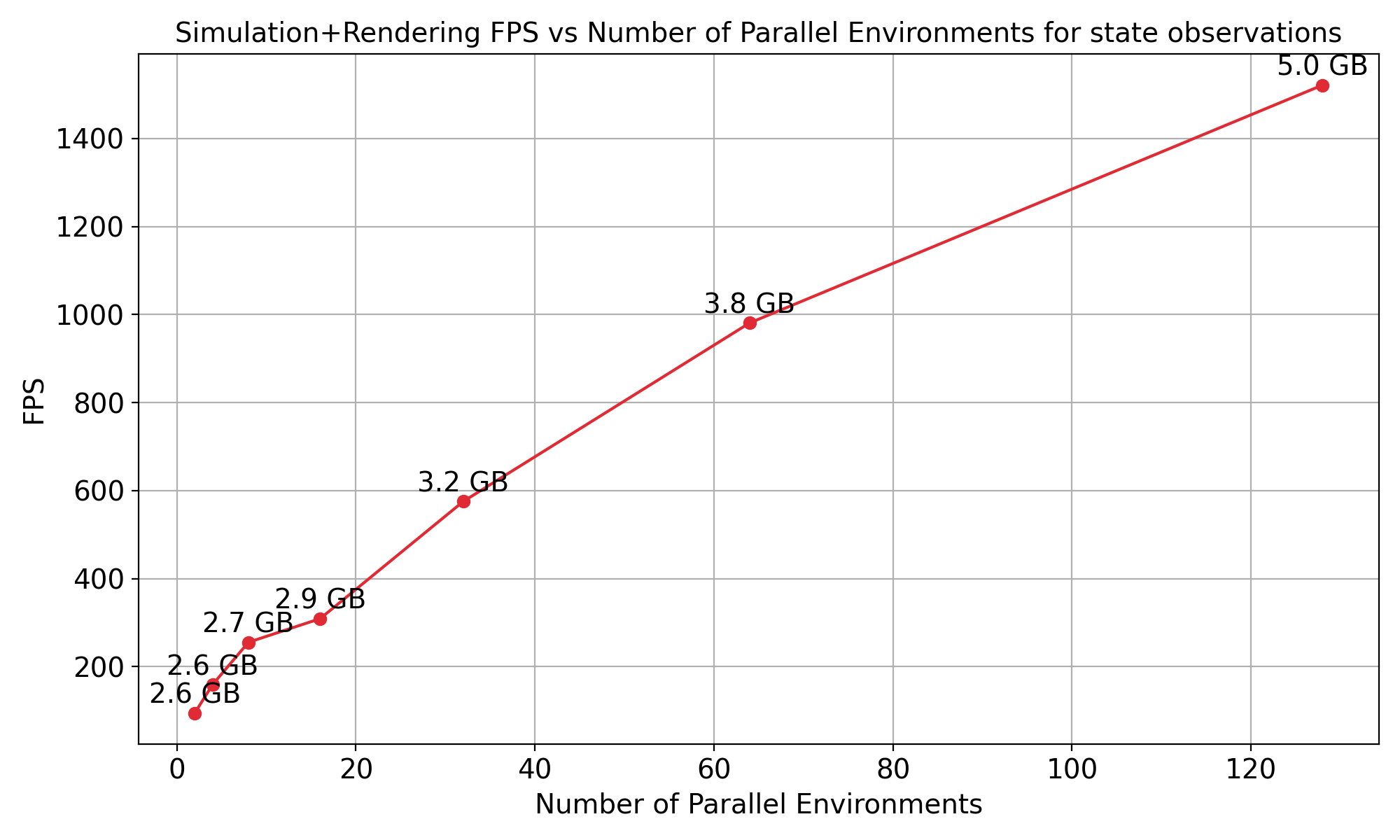}
        \caption{RoboBenchMart scene simulation performance showing FPS vs number of environments for state-only observations.}
        \label{fig:sup:perf_state}
    \label{fig:bench:pick-to-basket}
    \end{minipage}
    \hspace{1em}
    \begin{minipage}[t]{0.45\textwidth}
        \centering
        \includegraphics[width=\linewidth]{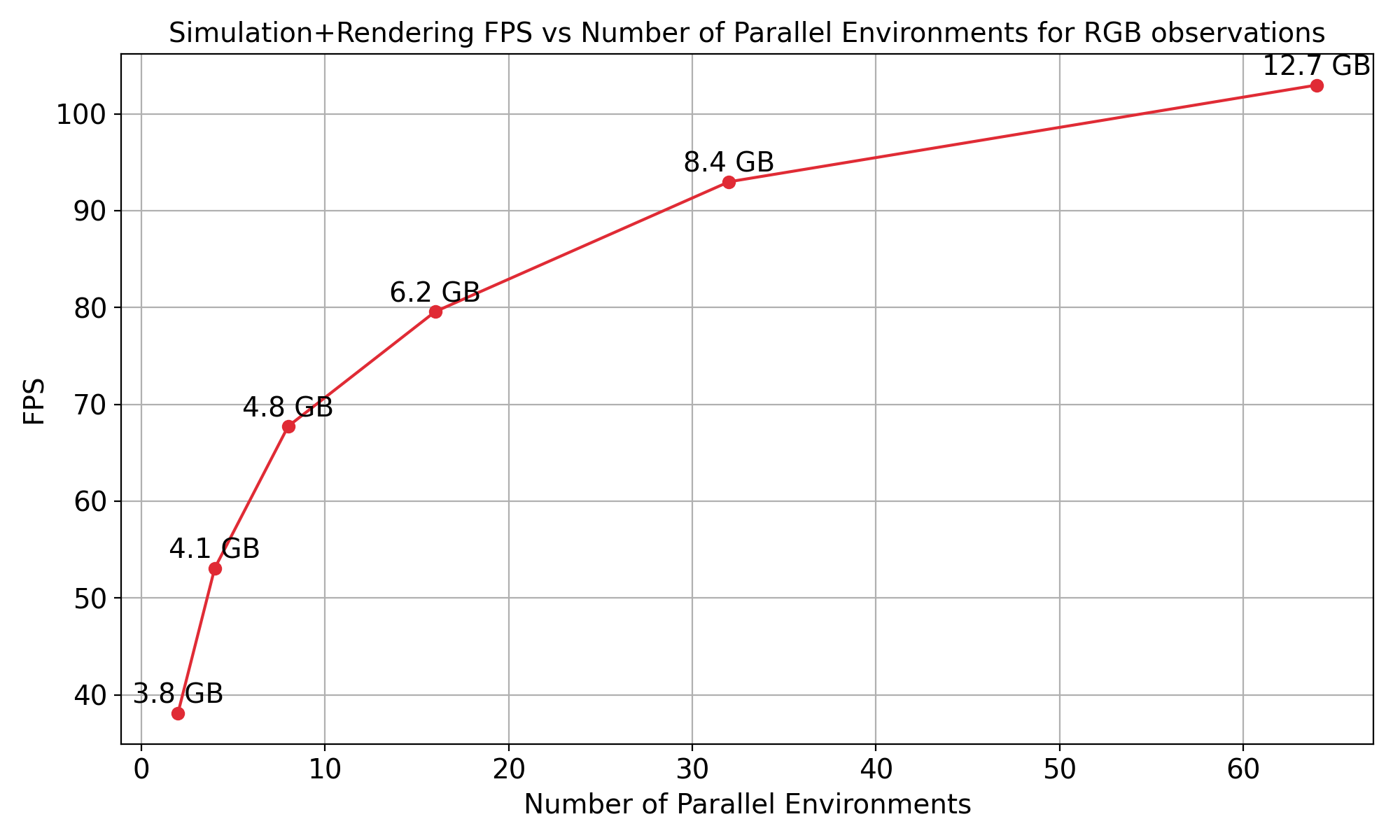}
        \caption{RoboBenchMart scene simulation performance showing FPS vs number of environments for realistic RGB camera observations.}
        \label{fig:sup:perf_rgb}
    \end{minipage}
    
\end{figure*}

In Figure~\ref{fig:sup:perf_state} we show performance of state-only setup, i.e. joint positions and velocities and extra task-specific information, e.g., goal position, end-effector position.
In Figure~\ref{fig:sup:perf_rgb} setup with rendering RGB sensor data from robot cameras is presented.
Experiments were conducted on Intel Xeon Gold 6278C CPU and NVIDIA V100 GPU.
We used default shader without ray tracing.

As it can be seen from Figures \ref{fig:sup:perf_state} and \ref{fig:sup:perf_rgb}, FPS grows non-linearly, the gain gradually decreases with increasing number of parallel environments due to the increasing number of assets.
Still, using state-only observations can help for training skill-specific RL policies, where a large number of parallel scenes is desired.

\section{Access and License}

\paragraph{Access} \RBM{} is publicly available on GitHub: \url{https://github.com/emb-ai/RoboBenchMart}

\paragraph{License} All assets are released under the CC BY-NC 4.0\footnote{https://creativecommons.org/licenses/by-nc/4.0/deed.en} license, and the codebase under the MIT\footnote{https://opensource.org/license/mit} license.

\FloatBarrier

\end{document}